\documentclass[10pt,twocolumn,letterpaper]{article}

\usepackage{iccv}              

\definecolor{mypink}{rgb}{.99,.90,.98}
\definecolor{lightgray}{HTML}{E6E6E6}
\definecolor{iccvblue}{rgb}{0.21,0.49,0.74}
\usepackage[pagebackref,breaklinks,colorlinks,allcolors=iccvblue]{hyperref}

\usepackage{colortbl}
\usepackage{multirow}
\usepackage{marvosym}       
\newcommand{\tinyskip}{\vspace{1mm}}

\def\paperID{1980} 
\def\confName{ICCV}
\def\confYear{2025}

\title{Towards Scalable Spatial Intelligence via 2D-to-3D Data Lifting}

\vspace{-2mm}
\author{
\fontsize{9.15}{11.0}\selectfont
Xingyu\,Miao$^{1}$\;\;\,Haoran\,Duan$^{3}$\;\;\,Quanhao\,Qian$^{2}$\;\;\,Jiuniu\,Wang$^{2}$\;\;\,Yang\,Long$^{1}$\;\;\,Ling\,Shao$^{4}$\;\;\,Deli\,Zhao$^{2}$\;\;\,Ran\,Xu$^{2}$\textsuperscript{\Letter}\;\;\,Gongjie\;Zhang$^{2}$\textsuperscript{\Letter}$^\dagger$
\tinyskip\\
{\fontsize{8.5}{11.0}\selectfont $^{1}$Durham University \quad\; $^{2}$DAMO\;Academy, Alibaba\;Group \quad\; $^{3}$Tsinghua University \quad\; $^{4}$UCAS-Terminus\;AI\;Lab} \\
{\footnotesize {\footnotesize \Letter}\,:\;Co-corresponding Author. \quad  $\dagger$: Project Lead. \;\;\,\quad\quad  \tt Project Page:\;\textcolor{red}{\href{https://ZhangGongjie.github.io/TowardsSSI-page/}{https://ZhangGongjie.github.io/TowardsSSI-page/}} }
}

\begin{document}
\twocolumn[{%
\renewcommand\twocolumn[1][]{#1}%
\maketitle

\vspace{-8mm}
\begin{center}
    \centering
    \vspace{-3mm}
    \captionsetup{type=figure}
    \includegraphics[width=0.965\linewidth]{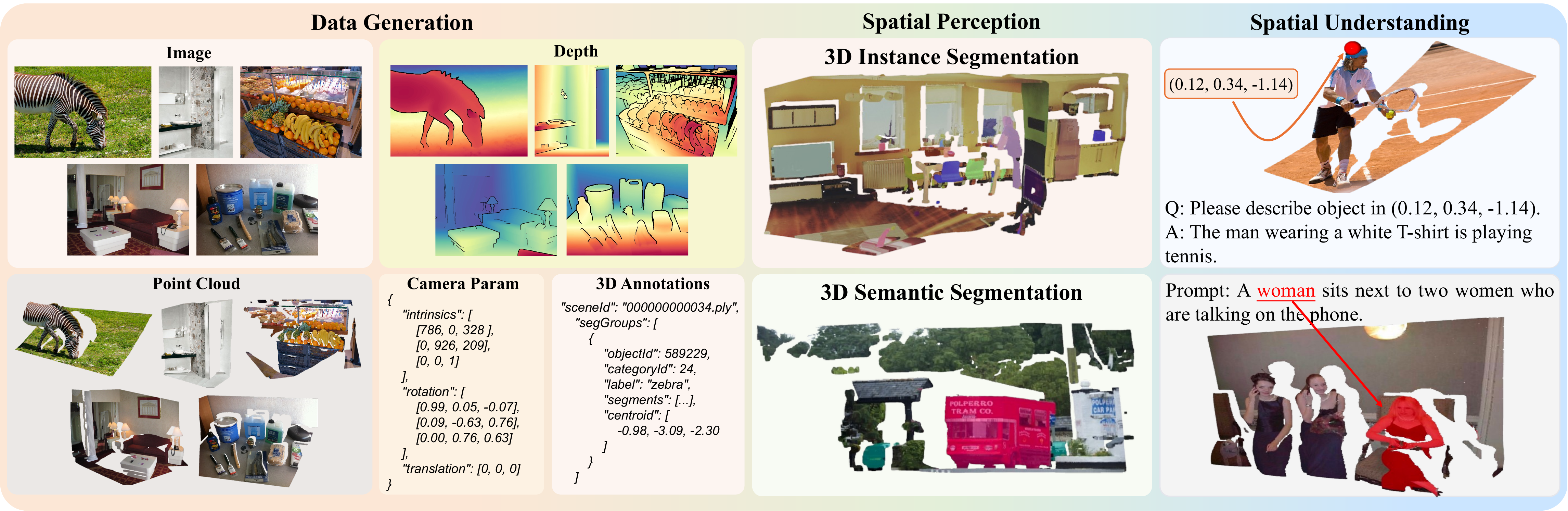}
    \vspace{-2.2mm}
    \captionof{figure}
    {\textbf{Scaling up spatial intelligence via 2D-to-3D data lifting.} Our pipeline mitigates the scarcity of spatial data by generating scale- and metric-authentic 3D data (point clouds, depth maps, camera poses, etc.) with rich annotations. Our generated data can support a wide range of tasks, including spatial perception and MLLM-based captioning, spatial reasoning, and grounding.}
    \vspace{+1.2mm}
    \label{fig:teaser}
\end{center}%
}]

\begin{abstract}
Spatial intelligence is emerging as a transformative frontier in AI, yet it remains constrained by the scarcity of large-scale 3D datasets. Unlike the abundant 2D imagery, acquiring 3D data typically requires specialized sensors and laborious annotation. In this work, we present a scalable pipeline that converts single-view images into comprehensive, scale- and appearance-realistic 3D representations — including point clouds, camera poses, depth maps, and pseudo-RGBD — via integrated depth estimation, camera calibration, and scale calibration. Our method bridges the gap between the vast repository of imagery and the increasing demand for spatial scene understanding. By automatically generating authentic, scale-aware 3D data from images, we significantly reduce data collection costs and open new avenues for advancing spatial intelligence. We release two generated spatial datasets, i.e., COCO-3D and Objects365-v2-3D, and demonstrate through extensive experiments that our generated data can benefit various 3D tasks, ranging from fundamental perception to MLLM-based reasoning. These results validate our pipeline as an effective solution for developing AI systems capable of perceiving, understanding, and interacting with physical environments.
\end{abstract}

\section{Introduction}
\label{sec:intro}
Spatial intelligence — the ability to perceive, reason about, and interact with 3D environments — is poised to drive the next wave of AI breakthroughs, with promising applications ranging from autonomous robotics \cite{fan2024integrating} to immersive AR/VR \cite{duan2022survey,miao2025rethinking}. Much like the success of multi-modal large language models (MLLMs), advancing spatial intelligence relies on the availability of massive, diverse, and annotated data. However, unlike the abundant text, images, or videos from the internet, acquiring spatial data demands specialized hardware (e.g., LiDAR) and labor-intensive, costly collection and labeling processes. This critical bottleneck has significantly limited the development of spatial intelligence, and the field's long-awaited ``ImageNet moment" remains out of reach.

Existing attempts to pursue scalable spatial data largely fall into three categories, yet each faces distinct drawbacks:
\vspace{-5mm}
\begin{itemize}
    \item \textbf{Simulation-based approaches:} Although simulation-based methods using game‐engine simulators (e.g., NVIDIA Isaac Gym \cite{makoviychuk2021isaac}) enable fast and cost‐effective data generation under controlled conditions, they often encounter a substantial sim‐to‐real gap \cite{tobin2017domain,peng2018sim,albardaner2024sim,nvidia2024}. This gap stems from the fact that the simplified geometric and physical models in simulation engines do not capture the full complexity and variability of real-world scenes. Thus, models trained entirely with simulation can struggle to generalize when exposed to the intricate and heterogeneous nature of real environments.

    \item \textbf{AI-generated 3D assets:} Although recent AI-based 3D generation can scale easily, current methods generally produce 3D assets limited to single objects \cite{poole2022dreamfusion,hong2023lrm,chung2023luciddreamer,lin2025diffsplat,lan2024ln3diff,he2024gvgen,hong20243dtopia,cao2024avatargo}. Scene-level generation remains challenging \cite{li2024dreamscene,bian2024dynamiccity,chung2023lucid,zhou2024dreamscene360,zhou2024holodreamer}. Generated scenes often exhibit disproportionate elements and unrealistic appearances. Texture and lighting renderings can hardly replicate natural conditions accurately, while complex scenes typically feature illogical object arrangements that deviate from real-world layouts. Moreover, most existing methods tend to produce cartoon-like 3D assets, limiting their applicability in realistic scenarios.
    
    \item \textbf{Sensor-captured data:} While offering high-fidelity authentic 3D data, data acquired via specialized hardwares (e.g., LiDAR and RGB-D camera) \cite{andersen2012kinect} incur high costs in both collection and annotation. These datasets are usually domain-specific (often indoors) and relatively small in scale, such as ScanNet \cite{dai2017scannet} (1,503 scenes) and Structured3D \cite{Structured3D} (3,500 scenes).
\end{itemize}
\vspace{-1.2mm}
On the other hand, 2D imagery datasets (e.g., COCO \cite{lin2014microsoft}, Objects365-v2 \cite{shao2019objects365}, OpenImages \cite{kuznetsova2020open}, etc.) encompass web-scale, richly annotated imagery across diverse scenes, objects, and tasks — thus fueling the success of MLLMs. However, their potential to advance spatial intelligence remains largely untapped.

To address the limitations of existing spatial datasets, we present a novel data-generation pipeline that \emph{lifts} large-scale 2D image datasets into high-quality, richly annotated, 3D representations covering diverse scenes and tasks (as shown in \cref{fig:teaser}). Rather than relying on simulation, purely generative methods, or specialized hardware-captured scenes, our approach leverages the rich visual content of 2D images to construct metric-scale 3D scenes. This ``2D-to-3D" pipeline effectively bridges the data gap in spatial intelligence by producing realistic, diverse environments at a fraction of the cost and complexity. Unlike simulation-based or AI-generated 3D data, our method preserves real-world textures and appearances; unlike sensor-captured data, it is not restricted by domain or hardware constraints and can be easily scaled up. By applying our pipeline to richly annotated 2D datasets such as COCO \cite{lin2014microsoft} and Objects365-v2 \cite{shao2019objects365}, we create \textit{COCO-3D} and \textit{Objects365-v2-3D}, marking the first large-scale expansion of spatial data to $\sim$2M distinct scenes spanning over 300 categories across diverse in-the-wild conditions, including indoor, outdoor, and mixed scenarios. These extensive 3D resources provide a robust foundation for training and evaluating spatial intelligence and embodied AI models across a wide range of tasks.

Extensive experiments show that our synthesized spatial data significantly enhance performance in various 3D perception tasks, including instance segmentation, semantic segmentation, and referring instance segmentation. Our results further indicate that tasks involving 3D LLMs—such as 3D dense captioning and 3D QA—also benefit from our data. Our results demonstrate that scalable 2D-to-3D lifting is a cost-effective and powerful strategy for advancing spatial intelligence and developing AI systems that truly understand and interact with the physical world. Enhanced quality and diversity in our 3D data directly translate to more accurate spatial perception, confirming the effectiveness of our data generation approach.

In summary, our contributions are threefold. \textit{First}, we propose a spatial data generation pipeline that constructs diverse and large-scale metric-scale 3D scenes from 2D images. \textit{Second}, we release large-scale spatial datasets — \textit{COCO-3D} and \textit{Objects365-v2-3D} — comprising $\sim$2M scenes and more than 300 categories across diverse environments. \textit{Third}, extensive experiments demonstrate that our generated data improve the performance of various spatial tasks including instance segmentation, semantic segmentation, referring instance segmentation, question answering, and dense captioning, validating that the proposed ``2D-to-3D'' can serve as a foundational paradigm for scalable spatial intelligence.

\section{Related Work}
\label{sec:related_work}

\begin{figure*}
    \centering
    \vspace{-2mm}
    \includegraphics[width=0.915\linewidth]{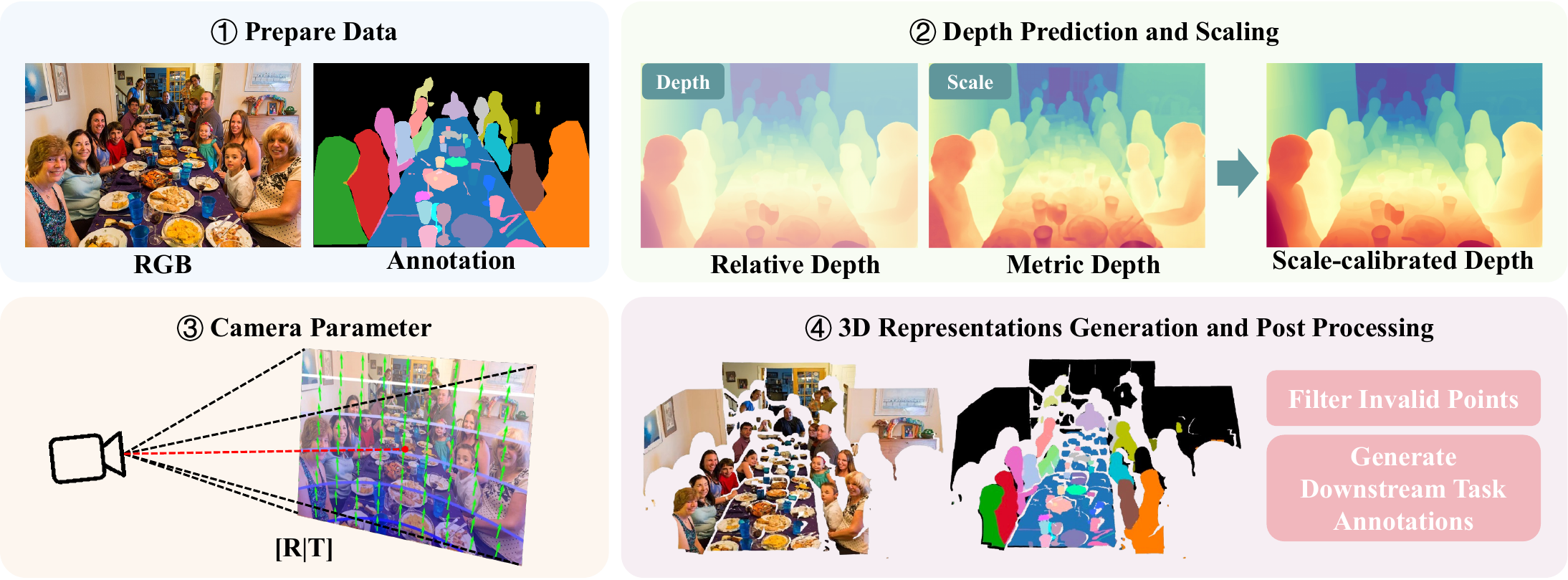}
    \vspace{-2mm}
    \caption{\textbf{Our proposed ``2D-to-3D data lifting'' pipeline.} First, we generate a scale-calibrated depth map by integrating scale-invariant and scale-aware depth estimation. Next, predicted camera parameters are used to project images into 3D space and remove invalid points. Finally, original 2D image annotations are also lifted to 3D, resulting in a fully annotated 3D representation for various downstream tasks.}
    \label{fig:pipline}
    \vspace{-2.5mm}
\end{figure*}

\noindent \textbf{Spatial Intelligence.\;}
In recent years, spatial intelligence has emerged as a critical frontier in computer vision and robotics. Early research focused on lightweight models for perception tasks \cite{schult2023mask3d,lai2023mask,metadetr,yin2024sai3d,lu2023query,huang2024openins3d,Luo_2021_ICCV,duan2023dynamic,zhang2022accelerating,zhang2024semantic,hua2025attention,luo2024modeling,zhang2023online,luo2023transpillars}, emphasizing efficient feature extraction and local detail capture. Building on these foundations, recent works has advanced toward object-level 3D MLLMs \cite{xu2024pointllm,guo2023point,wang2023beyond,qi2024gpt4point,liu20233daxiesprompts,qi2024shapellm,tang2024minigpt,NeurIPS24} and scene-level models that integrate visual and linguistic cues \cite{guo2023viewrefer,3dllm,yang2023llmgrounder,zhou2023uni3d,chen2024ll3da,yang2024thinking,huang2023embodied,yuan2024visual,huang2024chat,Chen_2024_CVPR,yang2023lidar,cheng2024spatialrgpt,linghu2025multi,zhu2024llava,man2024situation,zheng2024video3dllmlearningpositionaware,GPT4Scene}. These integrated approaches enable advanced spatial understanding and reasoning by leveraging natural language to interpret complex scenes. Moreover, vision-language-action (VLA) models—such as OpenVLA \cite{kim24openvla}, $\pi_0$ \cite{black2024pi_0}, RT-2 \cite{rt22023arxiv}, and Octo \cite{team2024octo}—demonstrate effective end-to-end mapping from perception to action via large-scale pre-training and fine-tuning, empowering robots to navigate complex scenes and perform sophisticated tasks.

\vspace{+1.5mm}
\noindent \textbf{3D Datasets.\;}
The scarcity of large-scale, diverse, and annotated 3D datasets remains a major bottleneck for advancing spatial intelligence. Existing approaches to obtaining 3D data generally fall into three categories. First, datasets generated by 3D simulator engines \cite{makoviychuk2021isaac,miceli2023dialogue,gulino2023waymax,liang2024aide,sun2024lidarf,haghighi2024taming} have been widely used. However, these methods often struggle when models trained in simulation are applied to real-world tasks. The simple geometric and physical models in simulation do not capture the detailed variability of actual environments, such as natural lighting changes, complex textures, and dynamic interactions. As a result, models trained entirely with simulated data tend to collapse when applied to real-world scenarios. Bridging the gap between simulation and reality remains an open question. Second, recent AI-based 3D asset generation methods \cite{chung2023luciddreamer,xu2024grm,xu2024instantmesh,ren2025l4gm,bahmani2024tc4d,chen2025meshxl,zhou2025diffgs,yu2024wonderworld,li2024sat2scene,zheng2024unified} have shown notable progress, but they still tend to produce cartoon-like outputs that lack the realism required for practical applications. Third, sensor-captured datasets using specialized hardware (e.g., LiDAR and RGB-D cameras) \cite{rozenberszki2022language,achlioptas2020referit3d,zhang2023multi3drefer,chen2020scanrefer,azuma2022scanqa,chen2021scan2cap} offer high-fidelity 3D representations of real-world scenes. However, their collection and annotation are costly, leading to relatively small-scale datasets that are predominantly confined to specific domains (e.g., indoor scenes) and often exclude dynamic or moving objects. Differently, our method fills the data gap in spatial intelligence by generating realistic and diverse environments at low cost. Unlike AI-generated or simulated data, our approach employs real appearances, ensuring high fidelity. We can lift any well-annotated 2D dataset, like COCO and Objects365-v2, to offer a solid basis for training and evaluating spatial intelligence across a wide range of tasks.

Two recent works have attempted to generate 3D data directly from 2D inputs for spatial tasks. SpatialRGPT \cite{cheng2024spatialrgpt} constructs 3D scene graphs for region-level question answering, focusing on basic spatial measurements (e.g., left-right, front-back, up-down) QA tasks. However, the 3D data it generates lack fine-grained details, limiting their applicability. SpatialBot \cite{cai2024spatialbot} fuses depth from ZoeDepth \cite{bhat2023zoedepth} with image features but omits camera intrinsics and extrinsics—leading to inconsistent results in small indoor spaces versus large outdoor environments. Consequently, both methods produce coarse or incomplete 3D data tailored primarily to specific tasks, often missing crucial fine-scale geometry. In contrast, our work systematically integrates both scale-invariant and scale-aware depth estimation along with camera parameters. This approach captures real-world scale and preserves fine-grained details, resulting in significantly higher-quality 3D data. Our novel pipeline supports a broader range of real-world spatial tasks and is validated by comprehensive experiments demonstrating its effectiveness.

\section{Data Generation and Statistics}
In this section, we first describe our data processing procedure for generating metric-scale 3D data. Then, we briefly discuss the statistics of the generated datasets.

\subsection{Data Processing and Generation Pipeline}
\begin{figure}
    \centering
    \includegraphics[width=\linewidth]{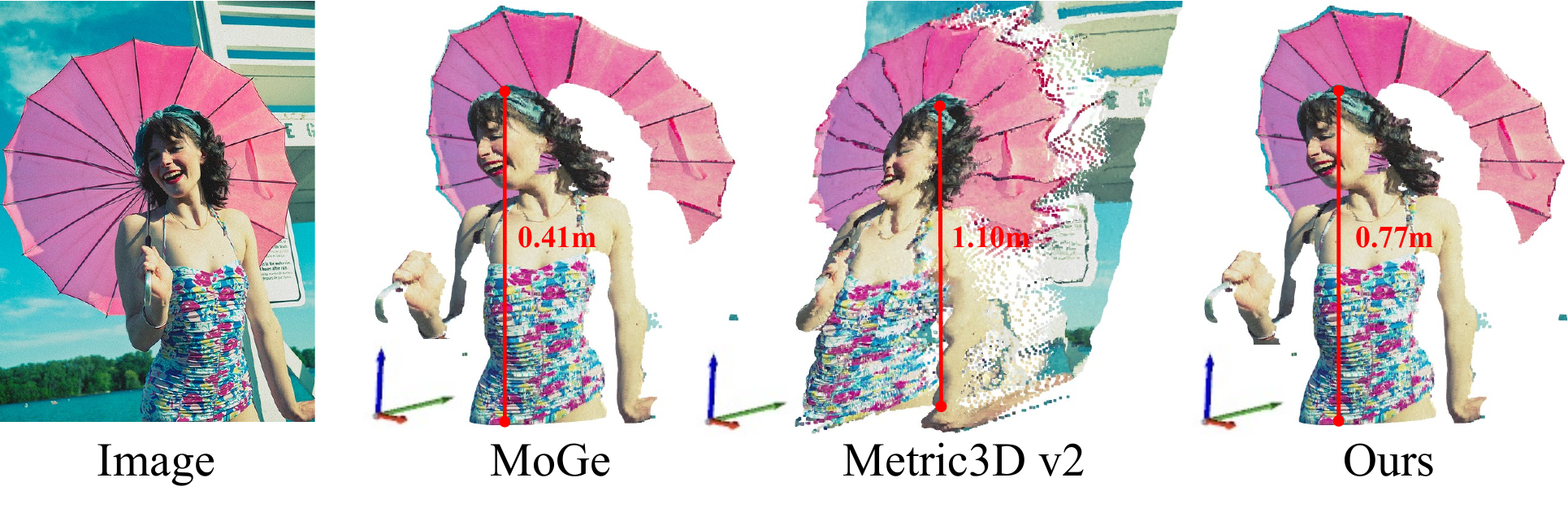}
    \vspace{-8mm}
    \caption{\textbf{Comparison of 3D representations produced by different depth estimation methods.}
    Our method generates 3D assets with correct scale, refined structure, and realistic appearance from RGB images.}
    \label{fig:scale}
    \vspace{-3mm}
\end{figure}

Our data pipeline automatically produces 3D, region-aware annotations from 2D images by building scale-calibrated 3D representations for each image, as shown in \cref{fig:pipline}. This is accomplished through four steps: \textit{i)} relative depth estimation, \textit{ii)} metric depth estimation, \textit{iii)} scale calibration, and \textit{iv)} camera parameter prediction for projecting 2D objects into 3D space. Relative depth estimation captures fine-grained geometry but lacks scale information, whereas metric depth estimation provides a precise global scale yet may trade off local geometry. By integrating these two approaches and calibrating the resulting depth, our method achieves 3D representations that capture both refined details and consistent real-world scale (See \cref{fig:scale}). Specifically, for each image we estimate both the relative-depth and metric-depth maps, compute a scaling factor over valid regions, and then apply it to the relative-depth map. We subsequently use predicted camera intrinsics and extrinsics to transform the scale-calibrated depth into a unified 3D coordinate system. To generate 3D annotations, we either project existing 2D pixel-level annotations into 3D using scale-calibrated depths and camera parameters, or first generate 2D annotations with an open-vocabulary detection/segmentation model and then lift them into the 3D space. Finally, we remove invalid points from the synthesized dataset and manually verify the resulting 3D point clouds along with their annotations.

\vspace{-4mm}
\paragraph{Relative Depth Estimation.}
Monocular depth estimation is widely regarded as an ill-posed problem, often yielding only relative depth predictions. Many approaches address this challenge by modeling depth in an affine-invariant manner, leveraging more diverse data sources \cite{ke2023repurposing, yang2024depthv2, yang2024depth}. However, such methods may struggle with capturing fine geometric details. To obtain relatively accurate relative depth maps with robust 3D geometry, we employ MoGe \cite{wang2024moge} as our relative depth predictor. We observe that MoGe \cite{wang2024moge} is robust for images captured in real-world conditions. Given a single-view image, MoGe \cite{wang2024moge} first estimates a 3D point cloud—providing a richer geometric representation—before deriving the relative depth map. Additionally, by using a multi-scale local geometric loss that penalizes local differences in the 3D point cloud under independent affine alignments, MoGe \cite{wang2024moge} achieves relatively accurate local geometric accuracy. However, while relative depth estimation performs well in geometric shape recovery, purely relative depth estimation lacks scale information, limiting precise distance interpretation in real environments.

\vspace{-4mm}
\paragraph{Metric Depth Estimation.}
To address this issue, we also need to determine the scale of the scene. Numerous works have explored methods for recovering metric depth from a single image \cite{bhat2021adabins,bhat2023zoedepth,eigen2014depth,hu2024metric3d,li2024binsformer, piccinelli2024unidepth, yin2023metric3d}, but resolving scale ambiguity remains a key challenge. A straightforward approach is to first generate a relative depth map, then fine-tune a metric depth head on datasets containing ground-truth depth. While this method may not capture fine-grained geometry perfectly, it generally provides the correct global scale. However, these methods often rely heavily on data from specific sensors, such as RGB-D cameras, LiDAR, or calibrated stereo cameras (e.g., KITTI \cite{geiger2013vision} or NYU \cite{SilbermanECCV12}). This dependency limits their applicability to particular scenarios and can lead to overfitting to the depth scales of the dataset and the camera, resulting in poor robustness when applied to wild images. Metric3D v2 \cite{hu2024metric3d} includes focal length as input and employs end-to-end training to predict both metric depth and surface normals. The model is trained jointly across a variety of indoor and outdoor scenes, which reduces the tendency to overfit the depth distributions of individual datasets. We find that Metric3D v2 \cite{hu2024metric3d} exhibits strong robustness on images captured in real-world environments. In most cases, combining Metric3D v2 \cite{hu2024metric3d} with camera intrinsics results in a reasonable scale. Hence, we use Metric3D v2 \cite{hu2024metric3d} as our metric depth estimation model. Despite jointly optimizing depth and normals, it still struggles to recover detailed 3D geometry in outdoor scenes, especially those involving people.

\vspace{-4mm}
\paragraph{Scale-calibrated Depth Map.}
Through the above relative depth estimation and metric depth estimation components, we can obtain the relative depth $d_r$ from MoGe \cite{wang2024moge} and metric depth $d_m$ from Metric3D v2 \cite{hu2024metric3d}. First, we combine the relative depth $d_r$ and the metric depth $d_m$ to determine a scaling factor. Subsequently, we scale the relative depth to obtain the scale-calibrated depth. To be more concrete, given an image, we generate relative depth and metric depth maps of identical dimensions. We begin by identifying and excluding invalid points, denoted by the set $\mathcal{I}$. Let $\mathcal{V}$ represent the set of valid points after exclusion, and $|\mathcal{V}|$ denote the number of valid points. The scaling factor $s$ is computed based on the average values of the valid relative and metric depths:
\begin{equation}
    s = \frac{\frac{1}{|\mathcal{V}|} \sum_{i \in \mathcal{V}} d_{m,i}}{\frac{1}{|\mathcal{V}|} \sum_{i \in \mathcal{V}} d_{r,i}}
    \label{eq:scale_factor}
\end{equation}
Using the scaling factor $s$, the scale-calibrated depth $d_{sc,i}$ for each valid point $i$ is obtained by scaling the relative depth $d_{r,i}$:
\begin{equation}
    d_{sc,i} = s \cdot d_{r,i}, \quad \forall i \in \mathcal{V}
    \label{eq:metric_depth}
\end{equation}

\vspace{-4mm}
\paragraph{Camera Parameter Prediction.}
When projecting 2D images into 3D space, precise camera parameters are crucial. These parameters include intrinsic parameters (focal length and principal point) and extrinsic parameters (the camera's position and orientation), which together define how the image aligns with real-world 3D structures. Since many in-the-wild images lack ground-truth camera parameters, we estimate them in two steps. First, we adopt WildCamera \cite{zhu2023tame} to predict the intrinsic parameters, leveraging its scale-awareness and cropping detection to accurately recover the 2D principal point and focal length. Next, we rely on PerspectiveFields to infer the extrinsic parameters (i.e., the camera’s pose relative to a typical 3D coordinate system). Specifically, PerspectiveFields provides per-pixel upward vectors and latitude values, allowing us to construct a rotation matrix that aligns the resulting point cloud with the standard 3D dataset orientation (z-axis upward), just like ScanNet canNet \cite{dai2017scannet} and Structured3D \cite{Structured3D}, thereby ensuring the reconstructed scene matches the real-world perspective.

\vspace{-4mm}
\paragraph{Constructing Scale-calibrated 3D Representations and 3D Annotations.}
We can further obtain the scale-calibrated depth $d_{sc,i}$, the camera intrinsic matrix $K$ from WildCamera \cite{zhu2023tame}, and camera extrinsic parameters $[R|T]$ from PerspectiveFields \cite{jin2023perspective}. Utilizing $K$ and $[R|T]$, we project each image into 3D space to generate a metric 3D point cloud. For each valid pixel with coordinates $(u_i, v_i)$, the projection is performed as follows:
\begin{equation}
    \mathbf{P}_i^{\text{cam}} = d_{sc,i} \cdot K^{-1} \begin{bmatrix} u_i \\ v_i \\ 1 \end{bmatrix}
    \label{eq:camera_projection}
\end{equation}
\begin{equation}
    \mathbf{P}_i^{\text{world}} = R \cdot \mathbf{P}_i^{\text{cam}} + T
    \label{eq:world_projection}
\end{equation}
Here, $\mathbf{P}_i^{\text{cam}}$ represents the 3D point in the camera coordinate system, and $\mathbf{P}_i^{\text{world}}$ denotes the corresponding point in the world coordinate system. For segmentation annotations, we align the projection with the RGB images to directly generate 3D annotations. For bounding box annotations, we first determine the maximum and minimum depth values within the region to construct a 3D bounding box, which is then converted into standardized 3D annotations. Notably, for COCO we directly use the provided segmentation masks, whereas for Objects365-v2 —which only provides bounding box annotations — we utilize SAM~\cite{kirillov2023segment} to generate the masks. Finally, we manually select some scenes to verify that the generated 3D annotations correctly align with the 3D point clouds.

\begin{figure}
    \centering
    \includegraphics[width=\linewidth]{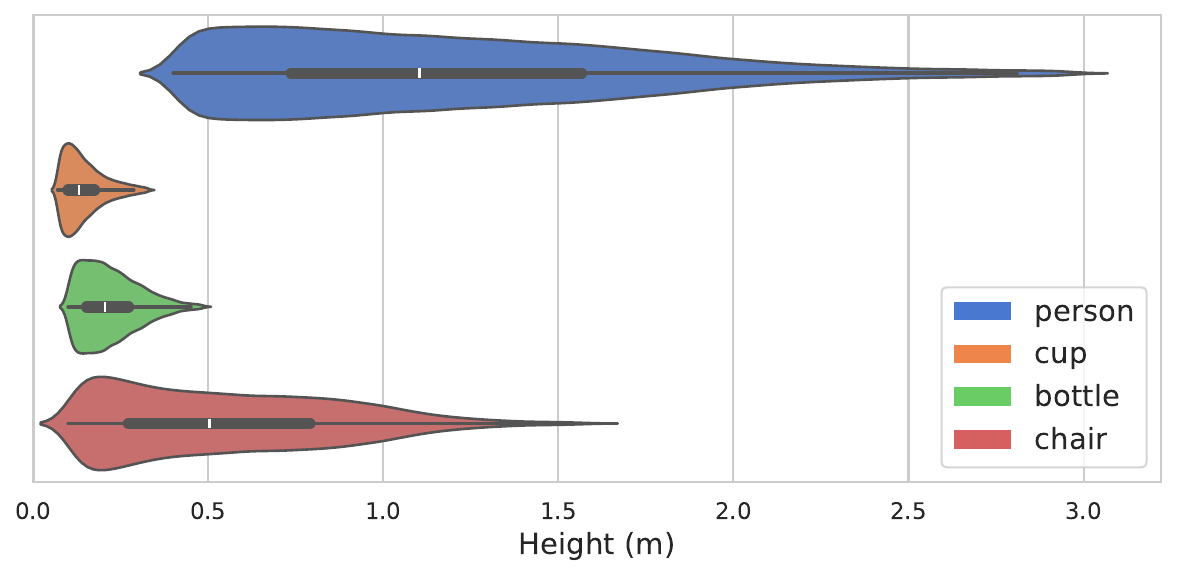}
    \vspace{-6.8mm}
    \caption{\textbf{Height distributions} for four common categories in COCO-3D. The height distribution of each category is reasonable and close to real-world measurements. Note that many objects appear shorter in the dataset due to truncation in images.}
    \label{fig:stat_height}
    \vspace{-1.5mm}
\end{figure}

\subsection{Dataset Statistics}
\label{sec:data_stat}
Based on the COCO dataset, we generate the COCO-3D dataset and use it as an example to demonstrate the statistics of our synthesized 3D data. Although COCO-3D is derived from COCO, we exclude regions with undefined geometry and remove invalid points, which leads to slight differences from the original COCO annotations. After filtering and validation, we end up with 117,183 generated 3D scenes in the training set and 4,951 3D scenes in the validation set. Notably, the scale of COCO-3D surpasses that of existing scene-level 3D datasets. (compared with 1,503 scenes in ScanNet \cite{dai2017scannet}, 3,500 scenes in Structured3D \cite{Structured3D}, and 90 scenes in Matterport3D \cite{Matterport3D}.) We ensure that every sample in both sets contains at least one valid object. The 3D scene data and their annotations will be made publicly available.

To further assess whether our synthesis process preserves real-world scaling, we analyze the height distributions of several categories with abundant instances (\emph{person, cup, bottle,} and \emph{chair}). Since many objects in COCO are partially captured or viewed from varying angles, direct measurements of object length and width are less straightforward. Therefore, we align each 3D scene with a world coordinate system and compute object heights. As shown in \cref{fig:stat_height}, these height distributions closely align with real-world expectations. For instance, the ``person" category ranges from about 0.5 m to 2.0 m to accommodate images that capture half-body views, seated or squatting individuals, and children. Note that many “person” instances appear shorter because only part of the body is visible. Other categories also fall within plausible ranges, further confirming the reliability of our synthetic dataset.

\section{Experiments}
We demonstrate the utility of our synthesized 3D data across a range of 3D tasks, such as instance segmentation, semantic segmentation, referring instance segmentation, dense captioning, and question answering.

\subsection{Experimental Settings}
\textbf{Data Setup.}
In this work, based on the COCO-3D data split described in \cref{sec:data_stat}, we conduct model training and performance evaluation on its training and test sets. For the ScanNet dataset, we adopt the commonly used split, selecting 1,201 scenes for training and 312 scenes for testing. For zero-shot perception evaluation, we test on the 312 ScanNet scenes, the ``Area\_5" scene of the S3DIS dataset, the test set of the Matterport3D dataset, and the test set of the Structured3D dataset. Note that, for Matterport3D, following the method in OpenRooms \cite{li2021openroomsendtoendopenframework}, we map the original categories to the 20 semantic classes corresponding to ScanNet and retain all other superset classes; for Structured3D, following Swin3D \cite{yang2023swin3dpretrainedtransformerbackbone}, we select 25 categories from the original 40—those whose frequency is greater than 0.001—for testing. For the referring instance segmentation task, we first filter the scenes to remove samples with excessively large ranges. Ultimately, we selected 36,619 references from the training set of RefCOCOg (Google) \cite{yu2016modeling} and 2,591 references from the evaluation set for testing. In addition, for the 3D LLM tasks, we first use LLaVA-ReCap (COCO-118K) from LLava-Next \cite{li2024llava} for pretraining. We then train on the training sets of both ScanRefer \cite{chen2020scanrefer} and Nr3D \cite{achlioptas2020referit_3d} for the 3D dense captioning task and evaluate on their respective test sets. For the 3D question answering task, we train on the ScanQA \cite{azuma2022scanqa} training set and evaluate on its test set.

\begin{table}
    \renewcommand{\arraystretch}{1.1}
    \setlength{\tabcolsep}{3pt}
    \centering
    \resizebox{\linewidth}{!}{%
        \begin{tabular}{c|ccc|ccc}
            \toprule
            Setting & Pretrain & Train & Val & mAP(\%) & mAP@0.25 & mAP@0.5 \\
            \cmidrule(r){1-1} 
            \cmidrule(lr){2-4} 
            \cmidrule(lr){5-7} 
             - & - & COCO-3D & COCO-3D & 22.95 & 42.82 & 33.96 \\
            - & - & ScanNet & ScanNet & 24.30 &  67.28 & 49.55 \\
            Pre-training & COCO-3D & ScanNet & ScanNet & 28.64 \textcolor{red}{{\scriptsize(+4.34)}} & 67.33 \textcolor{red}{{\scriptsize(+0.05)}} & 51.56 \textcolor{red}{{\scriptsize(+2.01)}} \\
            \bottomrule
        \end{tabular}
    }
    \vspace{-2.5mm}
    \caption{\textbf{Point cloud instance segmentation results}. Uni3D \cite{zhou2023uni3d}+Mask3D \cite{schult2023mask3d} is used as the baseline method. Pretraining on COCO-3D improves model performance.}
    \label{tab:ins-result}
    \vspace{-1.0mm}
\end{table}

\begin{table}
    \renewcommand{\arraystretch}{1.1}
    \setlength{\tabcolsep}{5pt}
    \centering
    \resizebox{\linewidth}{!}{%
        \begin{tabular}{c|c|c|ccc}
            \toprule
            Method & Val Data & Category & mAP(\%) & mAP@0.25 & mAP@0.5 \\ 
            \cmidrule(r){1-1} 
            \cmidrule(lr){2-2} 
            \cmidrule(lr){3-3} 
            \cmidrule(lr){4-6}
            \multirow{12}{*}{Uni3D \cite{zhou2023uni3d}} & 
            \multirow{8}{*}{ScanNet} & 
            Bed & 17.25 & 77.29 & 47.65 \\
            & & Chair & 38.39 & 82.99 & 64.24 \\
            & & Sofa & 29.29 & 78.59 & 55.20 \\
            & & Table & 16.49 & 37.78 & 48.30 \\
            & & Refrigerator & 15.21 & 37.96 & 44.03 \\
            & & Toilet & 60.27 & 91.49 & 87.07 \\
            & & Sink & 12.04 & 66.05 & 32.28 \\
            & &\cellcolor{mypink}{Avg.} & \cellcolor{mypink}{26.99} & 
            \cellcolor{mypink}{69.82} & \cellcolor{mypink}{51.74} \\ 

            \cmidrule(lr){2-2} 
            \cmidrule(lr){3-3} 
            \cmidrule(lr){4-6}
            &
            \multirow{4}{*}{S3DIS} & 
            Table & 8.57 & 29.99 & 12.53 \\
            & & Chair & 60.74 & 91.49 & 84.72 \\
            & & Sofa & 23.25 & 33.33 & 33.33 \\
            & &\cellcolor{mypink}{Avg.} & \cellcolor{mypink}{30.85} & 
            \cellcolor{mypink}{51.60} & \cellcolor{mypink}{43.53} \\ 
            \bottomrule
        \end{tabular}
    }
    \vspace{-2.5mm}
    \caption{\textbf{Point cloud instance segmentation zero-shot evaluation.} A model pre-trained on COCO-3D can directly generalize to sensor-captured 3D perception datasets such as ScanNet and S3DIS. Only overlapping categories are evaluated.}
    \label{tab:ins-zero}
    \vspace{-1.5mm}
\end{table}

\vspace{-4mm}
\paragraph{One Unified Hyperparameter Setting Across All Models and Datasets.} Hyperparameters and tricks such as voxelization grid size, number of points, learning rates, and point sampling strategies strongly influence performance in 3D perception tasks; however, their optimal values can vary substantially across datasets due to differences in scene structure and object distributions. For instance, hyperparameters fine-tuned for S3DIS may yield suboptimal results on ScanNet, and vice versa. Our synthetic COCO-3D dataset further magnifies these discrepancies as it includes both indoor and outdoor scenes with scales ranging from 1 meter to several hundred meters—well beyond the ~10-meter range common in ScanNet. Consequently, individually optimized hyperparameters for each dataset can introduce bias and fail to capture the diversity of real-world challenges.
Our ultimate goal is to develop \emph{scalable and generalizable} spatial intelligence — an objective that conflicts with the prevailing practice of tailoring hyperparameters meticulously for each dataset. To ensure fairness and to highlight the utility of our synthesized 3D data rather than hyperparameter tuning, we adopt \emph{a single, unified hyperparameter setting for all models and datasets, without any meticulously designed tricks}. This approach not only enhances reproducibility but also offers a clear and consistent baseline for assessing how our synthetic data can benefit real-world tasks.

\begin{table}
    \renewcommand{\arraystretch}{1.1}
    \setlength{\tabcolsep}{3pt}
    \centering
    \resizebox{\linewidth}{!}{%
    \begin{tabular}{l|ccc|ccc}
        \toprule
        Method & Pretrain & Train & Val & mIoU(\%) & mAcc & allAcc \\
        \midrule
        \multirow{3}{*}{SpUNet \cite{spconv2022}} 
            & -      & COCO-3D & COCO-3D & 20.02 & 26.70 & 77.59 \\
            & -      & ScanNet & ScanNet &     31.09  &   36.54   &   68.63    \\
            & COCO-3D & ScanNet & ScanNet & 62.48 \textcolor{red}{{\scriptsize(+31.39)}} 
            & 70.38 \textcolor{red}{{\scriptsize(+33.84)}} 
            & 84.89 \textcolor{red}{{\scriptsize(+16.26)}} \\
        \midrule
        \multirow{3}{*}{PTv2 \cite{wu2022point}} 
            & -      & COCO-3D & COCO-3D & 26.88 & 38.21 & 80.84 \\
            & -      & ScanNet & ScanNet &      51.04 &   58.73    &    78.17   \\
            & COCO-3D & ScanNet & ScanNet & 55.81 \textcolor{red}{{\scriptsize(+4.77)}} 
            & 63.19 \textcolor{red}{{\scriptsize(+4.46)}} 
            & 80.62 \textcolor{red}{{\scriptsize(+2.45)}} \\
        \midrule
        \multirow{3}{*}{Uni3D \cite{zhou2023uni3d}} 
            & -      & COCO-3D & COCO-3D & 38.16 & 50.14 & 84.00 \\
            & -      & ScanNet & ScanNet & 52.14 & 59.06 & 79.05   \\
            & COCO-3D & ScanNet & ScanNet & 55.83 \textcolor{red}{{\scriptsize(+3.69)}} 
            & 66.10 \textcolor{red}{{\scriptsize(+7.04)}} 
            & 81.31 \textcolor{red}{{\scriptsize(+2.26)}} \\
        \bottomrule
    \end{tabular}%
    }
    \vspace{-2.3mm}
    \caption{\textbf{Point cloud semantic segmentation results.} Pre-training on COCO-3D improves performance on ScanNet across multiple baseline methods.}
    \label{tab:sem-result}
\end{table}

\begin{table}
    \centering
    \scriptsize
    \renewcommand{\arraystretch}{1.3}
    \setlength{\tabcolsep}{5pt}
    \resizebox{\linewidth}{!}{%
        \begin{tabular}{c|cc|cc|cc|cc}
            \toprule
            \multirow{2}{*}{Category} 
            & \multicolumn{2}{c}{ScanNet \cite{dai2017scannet}} 
            & \multicolumn{2}{c}{S3DIS \cite{armeni20163d}} 
            & \multicolumn{2}{c}{Matterport3D \cite{Matterport3D}} 
            & \multicolumn{2}{c}{Structured3D \cite{Structured3D}}\\
            \cmidrule(lr){2-3}\cmidrule(lr){4-5}\cmidrule(lr){6-7}\cmidrule(lr){8-9}
             & IoU (\%) & Acc (\%) & IoU (\%) & Acc (\%) & IoU (\%) & Acc (\%) & IoU (\%) & Acc (\%) \\
            \midrule
            Wall         & 35.19 & 98.57 & 30.89 & 99.68 & 37.78 & 99.24 & 37.23 & 99.68 \\
            Bed          & 43.96 & 45.05 & -   & -   & 66.26 & 77.82 & 66.29 & 70.76 \\
            Chair        & 64.75 & 72.12 & 59.08 & 70.17 & 46.31 & 56.51 & 21.58 & 57.27 \\
            Sofa         & 51.77 & 59.85 & 5.29  & 5.36  & 41.78 & 47.32 & 45.09 & 49.34 \\
            Table        & 31.38 & 32.37 & 17.93 & 18.30 & 6.97  & 7.06  & 23.03 & 24.86 \\
            Refrigerator & 20.25 & 36.46 & -   & -   & 9.81  & 27.05 & 36.20 & 45.09 \\
            Toilet       & 43.63 & 71.02 & -   & -   & 43.63 & 71.02 & -   & -   \\
            Sink         & 39.67 & 48.13 & -   & -   & 39.67 & 48.13 & 9.50  & 48.98 \\
            Bookcase     & -   & -   & 18.23 & 19.73 & -   & -   & -   & -   \\
            Television   & -   & -   & -   & -   & -   & -   & 31.47 & 39.09 \\
            \midrule
            \cellcolor{mypink}{Avg.}         & \cellcolor{mypink}{41.33} & \cellcolor{mypink}{57.95} & \cellcolor{mypink}{26.28} & \cellcolor{mypink}{42.65} & \cellcolor{mypink}{45.91} & \cellcolor{mypink}{66.67} & \cellcolor{mypink}{33.80} & \cellcolor{mypink}{54.38} \\
            \bottomrule
        \end{tabular}
    }
    \vspace{-2.3mm}
    \caption{\textbf{Point cloud semantic segmentation zero-shot evaluation.} Models pre-trained on COCO-3D can directly generalize to sensor-captured 3D datasets including ScanNet, S3DIS, Matterport3D, and Structured3D. Only overlapping classes are evaluated.}
    \label{tab:sem-zero}
\end{table}

\vspace{-4mm}
\paragraph{Implementation Details.}
For 3D perception tasks, all experiments are carried out using Pointcept \cite{pointcept2023}. For 3D instance segmentation, we employ Uni3D \cite{zhou2023uni3d} as the backbone to extract per-point features and use the Mask3D \cite{schult2023mask3d} instance head. For 3D semantic segmentation, we use SpUNet \cite{spconv2022}, PTv2 \cite{wu2022point}, and Uni3D \cite{zhou2023uni3d} as the backbone for feature extraction and apply a two-layer MLP as the semantic head. For referring instance segmentation, experiments are performed with the official TGNN implementation \cite{huang2021text}. For 3D question answering and dense captioning tasks, we use the official LL3DA implementation \cite{chen2023ll3da}.

\begin{figure}
    \centering
    \includegraphics[width=\linewidth]{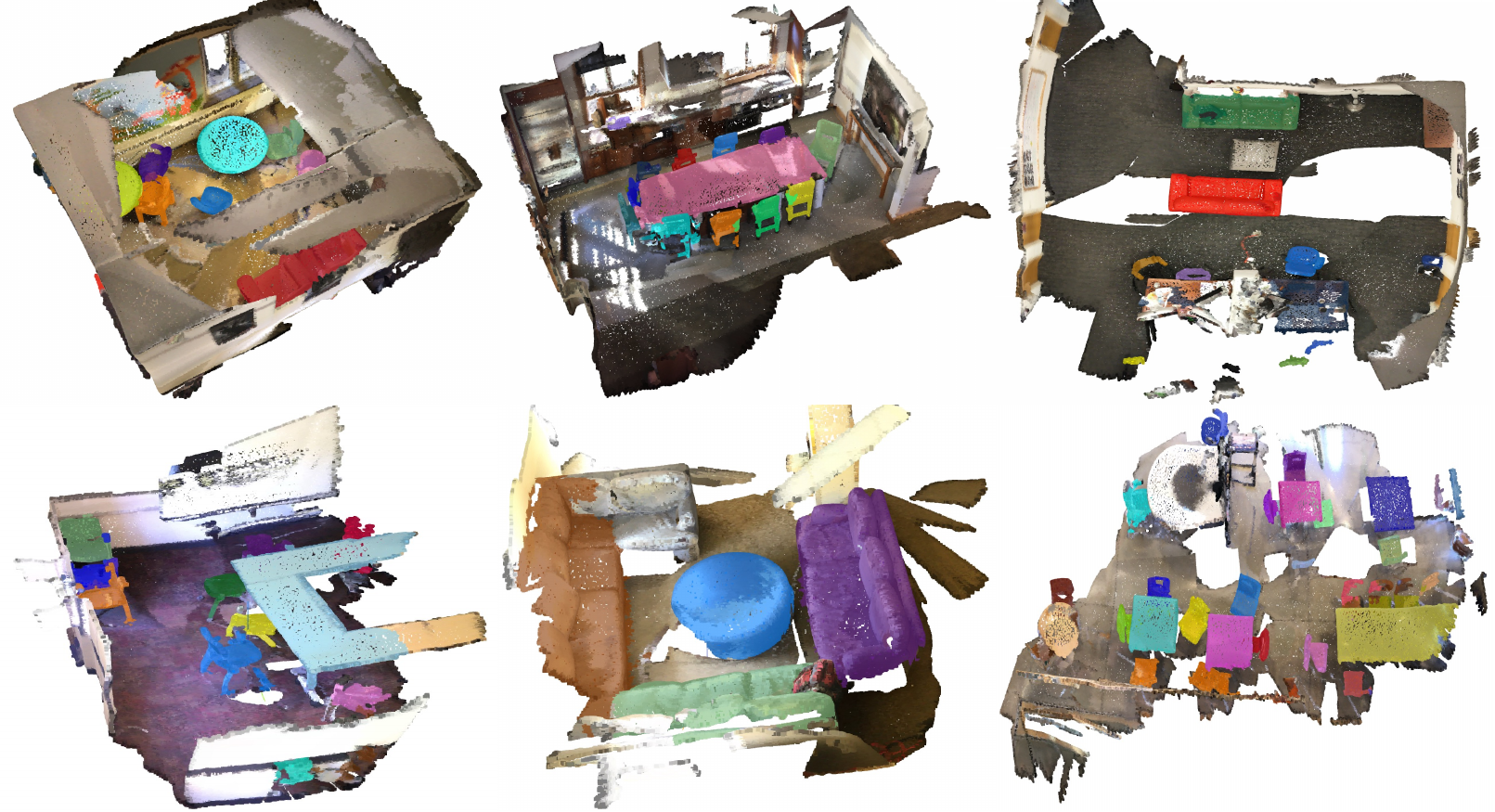}
    \vspace{-6mm}
    \caption{\textbf{Visualization of zero-shot point cloud instance segmentation results.} Despite significant differences between synthetic and real data, models trained on COCO-3D can directly generalize to ScanNet.}
    \label{fig:zero-ins-vis}
\end{figure}

\begin{table*}
    \renewcommand{\arraystretch}{1.3}
    \setlength{\tabcolsep}{5pt}
    \centering
    \vspace{-2.5mm}
    \resizebox{\linewidth}{!}{%
        \begin{tabular}{cc|cccc|cccc|cccc}
            \toprule
            \multirow{2}{*}{Pretrain data} & \multirow{2}{*}{Training Data} & \multicolumn{4}{c|}{ScanRefer} & \multicolumn{4}{c|}{Nr3D} & \multicolumn{4}{c}{ScanQA} \\
            \cmidrule(r){3-6} \cmidrule(lr){7-10} \cmidrule(lr){11-14}
             &  & C@0.5$\uparrow$ & B-4@0.5$\uparrow$ & M@0.5$\uparrow$ & R@0.5$\uparrow$ & C@0.5$\uparrow$ & B-4@0.5$\uparrow$ & M@0.5$\uparrow$ & R@0.5$\uparrow$ & C$\uparrow$ & B-4$\uparrow$ & M$\uparrow$ & R$\uparrow$ \\
            \cmidrule(r){1-2}\cmidrule(r){3-6}\cmidrule(lr){7-10}\cmidrule(lr){11-14}
            - & ScanRefer, Nr3D, ScanQA & 62.98 & 35.97 & 25.66 & 54.65 & 23.94 & 13.37 & 22.31 & 45.78 & 75.67 & 13.33 & 15.37 & 37.02 \\
            \cellcolor{mypink}COCO-3D & \cellcolor{mypink}ScanRefer, Nr3D, ScanQA & \cellcolor{mypink}67.04 & \cellcolor{mypink}36.74 & \cellcolor{mypink}26.18 & \cellcolor{mypink}54.82 & \cellcolor{mypink}25.20 & \cellcolor{mypink}15.28 & \cellcolor{mypink}23.13 & \cellcolor{mypink}47.01 & \cellcolor{mypink}79.11 & \cellcolor{mypink}14.14 & \cellcolor{mypink}15.99 & \cellcolor{mypink}38.31 \\
            & & 
            \textcolor{red}{\small(+4.06)} & \textcolor{red}{\small(+0.77)} & \textcolor{red}{\small(+0.52)} & \textcolor{red}{\small(+0.17)} & 
            \textcolor{red}{\small(+1.26)} & \textcolor{red}{\small(+1.91)} & \textcolor{red}{\small(+0.82)} & \textcolor{red}{\small(+1.23)} & 
            \textcolor{red}{\small(+3.44)} & \textcolor{red}{\small(+0.81)} & \textcolor{red}{\small(+0.62)} & \textcolor{red}{\small(+1.29)} \\
            \bottomrule
        \end{tabular}
    }
    \vspace{-2mm}
    \caption{
    \textbf{Performance of the LL3DA “3D generalist model" with or without pretraining on COCO-3D.} Pretraining on COCO-3D consistently boosts performance across all tasks. Note that the model is not specifically fine-tuned for each individual dataset.}
    \label{tab:ll3da_gen}
    \vspace{-2.5mm}
\end{table*}

\subsection{3D Instance Segmentation}
We evaluate the effectiveness of 3D synthetic data for improving 3D instance segmentation and design experiments to investigate whether combining the synthetic dataset COCO-3D with the real-world dataset ScanNet can enhance model training. The same training hyperparameters are applied to both datasets. As shown in \cref{tab:ins-result}, pre-training on COCO-3D followed by fine-tuning on ScanNet yields an improvement of approximately 4\% compared to training solely on ScanNet. From this comparison, we observe that synthetic data, despite differing from real-world distributions, can effectively compensate for the limitations of real data and thus provide an extra performance increase.

Surprisingly, even though our synthetic 3D data only captures partial view point clouds, it can still generalize effectively to complete view datasets like ScanNet \cite{dai2017scannet}. For instance, in zero-shot evaluations, the "Toilet" category reaches an mAP over 60\%, as presented in \cref{tab:ins-zero} and \cref{fig:zero-ins-vis}. However, due to discrepancies in class definitions, we only report the categories that overlap between COCO and ScanNet in \cref{tab:ins-zero} for zero-shot generalization evaluation.   


\begin{table}
    \renewcommand{\arraystretch}{1.1}
    \setlength{\tabcolsep}{3pt}
    \centering
    \resizebox{\linewidth}{!}{%
        \begin{tabular}{c|ccc|ccc}
        \toprule
        Setting & Pretrain & Train & Val & mIoU (\%) & Acc@0.25 & Acc@0.5 \\
        \cmidrule(r){1-1} 
        \cmidrule(lr){2-4} 
        \cmidrule(lr){5-7} 
              -     & -      & COCO-3D          & COCO-3D         & 19.33     & 27.34    & 17.32   \\ 
        Zero Shot & -  & COCO-3D          & ScanNet        & 10.10     & 13.26    & 10.92   \\
            \cmidrule(r){1-1} 
        \cmidrule(lr){2-4} 
        \cmidrule(lr){5-7}
        -     & -     & ScanNet & ScanNet & 26.10     & 35.0     & 29.00   \\
            Pre-training & COCO-3D & ScanNet & ScanNet & \textbf{32.47} \textcolor{red}{{\scriptsize(+6.37)}}     & \textbf{43.24} \textcolor{red}{{\scriptsize(+8.24\%)}}    & \textbf{37.12} \textcolor{red}{{\scriptsize(+8.12\%)}} \\    
        \bottomrule
        \end{tabular}
    }
    \vspace{-2.3mm}
    \caption{\textbf{Referring point cloud instance segmentation results. }TGNN \cite{huang2021text} pre-trained on COCO-3D can directly generalize to ScanNet in a zero-shot manner, and fine-tuning on ScanNet further boosts performance. }
    \label{tab:refering}
    \vspace{-4.5mm}
\end{table}

\subsection{3D Semantic Segmentation}
We investigate the impact of our 3D synthetic data on 3D semantic segmentation. \cref{tab:sem-result} presents results from three methods—SpUNet \cite{spconv2022}, PTv2 \cite{wu2022point}, and Uni3D \cite{zhou2023uni3d}—trained under varying strategies. We note that for SpUNet on ScanNet, direct training produces only moderate results, whereas pre-training on COCO-3D increases the overall accuracy by over 30\%, with commensurate increases in mIoU and mAcc. PTv2 and Uni3D likewise benefit significantly from synthetic-data pre-training, validating that COCO-3D imparts valuable prior knowledge. By comparing these outcomes, we find that leveraging synthetic data fosters stronger generalization to real scenes, allowing the model to learn more robust representations that improve segmentation quality across various categories.

\begin{figure}
    \centering
    \includegraphics[width=\linewidth]{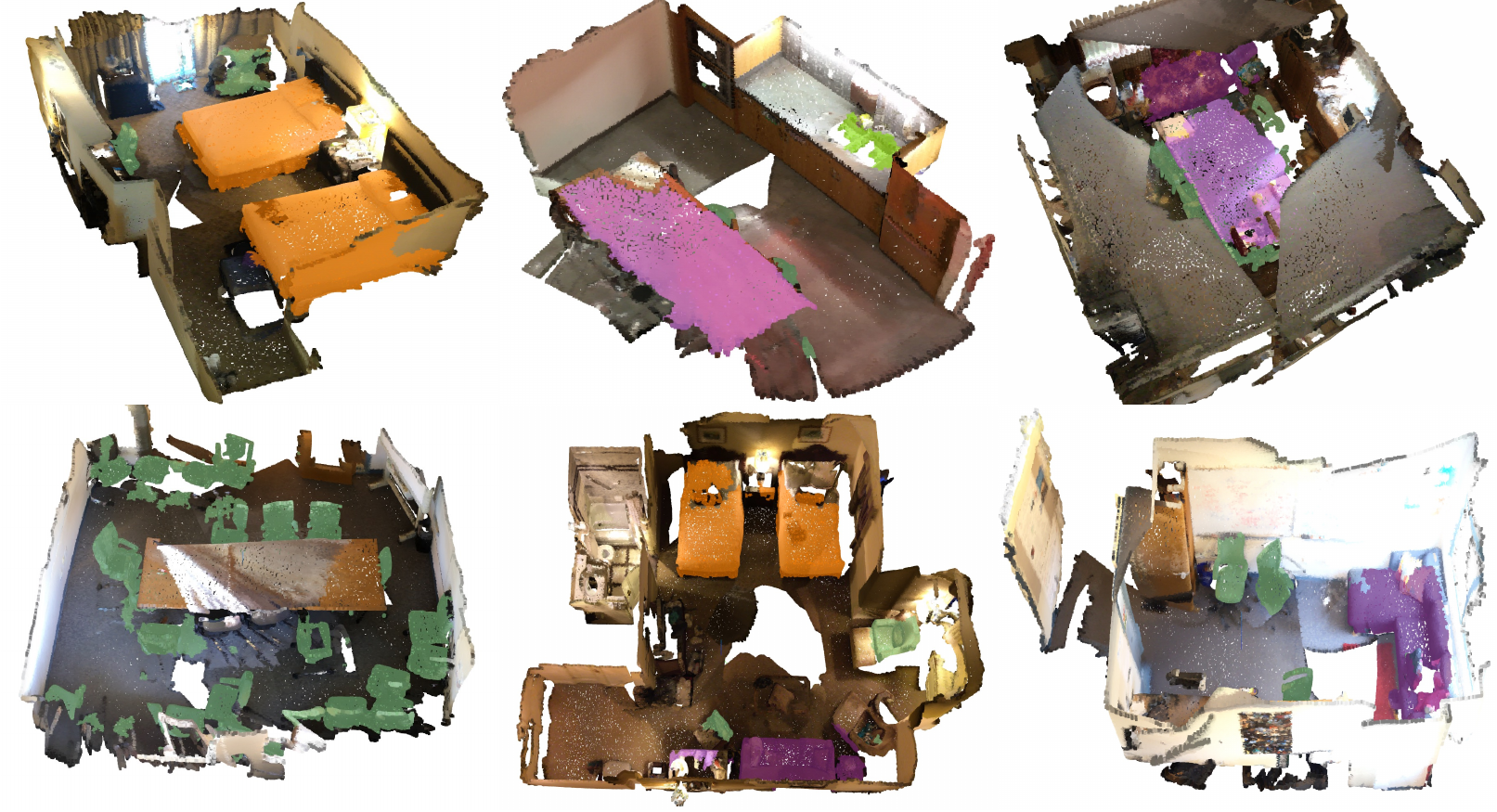}
    \vspace{-6.8mm}
    \caption{\textbf{Visualization of zero-shot point cloud semantic segmentation results.} Despite significant differences between synthetic and real data, models trained on COCO-3D can directly generalize to ScanNet.}
    \label{fig:zero-sem-vis}
    \vspace{-4.5mm}
\end{figure}

We also investigate zero-shot performance. In \cref{tab:sem-zero}, the models are trained solely on COCO-3D and then evaluated directly on ScanNet, S3DIS, Matterport3D, and Structured3D. The results demonstrate that synthetic data generalizes well to real-world datasets, enabling the model to segment multiple object categories with viable accuracy even without exposure to real data during training. This underscores the capability of COCO-3D to provide domain-relevant features for 3D semantic segmentation. Furthermore, visualization results in \cref{fig:zero-sem-vis} (for Uni3D) show that the model can reliably distinguish object classes under zero-shot conditions, reinforcing the insight that synthetic pre-training effectively supports semantic segmentation in real indoor environments.

\begin{table}
    \renewcommand{\arraystretch}{1.3}
    \setlength{\tabcolsep}{5pt}
    \centering
    \resizebox{\linewidth}{!}{%
        \begin{tabular}{l c c c c c}
            \toprule
             & \multirow{2}{*}{Base LLM} & \multicolumn{4}{c}{ScanQA (val)} \\
            \cmidrule(r){3-6}
             &         & C$\uparrow$ & B-4$\uparrow$ & M$\uparrow$ & R$\uparrow$ \\
            \midrule
            \multicolumn{6}{l}{\cellcolor{lightgray}\textit{\textbf{Task-specific models}}} \\
            ScanRefer+MCAN \cite{yu2019deep} & - & 55.4 & 7.9  & 11.5 & - \\
            ScanQA \cite{azuma_2022_CVPR} & BERT \cite{devlin2018bert} & 64.9 & 10.1 & 13.1 & 33.3 \\
            3D-VisTA \cite{zhu20233d} & - & 69.6 & 10.4 & 13.9 & 35.7 \\
            \midrule
            \multicolumn{6}{l}{\cellcolor{lightgray}\textit{\textbf{Zero-shot 2D LMMs}}} \\
            VideoChat2 \cite{li2024mvbench} & - & 49.2 & 9.6  & 9.5  & 28.2 \\
            LLaVA-NeXT-Video \cite{li2024llava} & Qwen2-7B \cite{qwen2} & 46.2 & 9.8  & 9.1  & 27.8 \\
            GPT-4V              & - & 59.6 & -    & 13.5 & 33.4 \\
            Gemini              & - & 68.3 & -    & 11.3 & 35.4 \\
            Claude              & - & 57.7 & -    & 10.0 & 29.3 \\
            \midrule
            \multicolumn{6}{l}{\cellcolor{lightgray}\textit{\textbf{Task-specific fine-tuned 3D LMMs}}} \\
            Scene-LLM \cite{fu2024scene} & - & 80.0   & 12.0 & 16.6 & 40.0 \\
            3D-LLM \cite{3dllm} & BLIP2-flan-t5 \cite{li2023blip} & 69.4 & 12.0 & 14.5 & 35.7 \\
            Chat-3D v2 \cite{huang2023chat} & Vicuna-7B \cite{zheng2024judging} & 87.6 & 14.0 & -    & - \\
            LEO \cite{huang2023embodied} & Vicuna-7B \cite{zheng2024judging} & \textbf{101.4} & 13.2 & 20.0 & 49.2 \\
            LLaVA-3D \cite{zhu2024llava} & Vicuna-7B \cite{zheng2024judging} & 91.7 & 14.5 & \textbf{20.7} & \textbf{50.1}\\
            \midrule
            LL3DA \cite{chen2023ll3da} & OPT-1.3B \cite{zhang2023opt} & 76.8 & 13.5 & 15.9 & 37.3 \\
            \cellcolor{mypink}LL3DA$^*$ \cite{chen2023ll3da} & \cellcolor{mypink}OPT-1.3B \cite{zhang2023opt} & \cellcolor{mypink}85.2 & \cellcolor{mypink}\textbf{15.8} & \cellcolor{mypink}16.9 & \cellcolor{mypink}40.5 \\
            & & \textcolor{red}{\small(+8.4)} &  \textcolor{red}{\small(+2.3)} & \textcolor{red}{\small(+1.0)} & \textcolor{red}{\small(+3.2)} \\
            \bottomrule
        \end{tabular}
    }
    \vspace{-2.2mm}
    \caption{
    \textbf{3D QA task results on ScanQA.} “*” denotes using COCO-3D for pretraining, which further boosts the 1.3B baseline performance and can match or even surpass 7B-parameter models fine-tuned specifically on this task.}
    \label{tab:ll3da_qa}
    \vspace{-3.5mm}
\end{table}

\subsection{3D Referring Instance Segmentation}
We evaluate the effectiveness of 3D synthetic data in enhancing the performance of existing baseline methods for the referring 3D instance segmentation task, which aims to accurately segment target instances within 3D scenes based on given referring expressions. To investigate whether the synthetic dataset COCO-3D can improve model training performance in tandem with the real dataset ScanNet, we design multiple experiments. We employ TGNN \cite{huang2021text} trained on ScanNet as the baseline. However, in zero-shot testing—where the model is trained on COCO-3D and directly tested on ScanNet—performance drops noticeably, which indicates the difficulty of transferring synthetic data without fine-tuning. This decline primarily stems from substantial differences between COCO-3D and ScanNet: \textit{i)} COCO-3D includes both indoor and outdoor scenes, whereas ScanNet focuses solely on indoor environments, which limits generalization; \textit{ii)} COCO-3D contains a larger proportion of persons, while ScanNet primarily labels indoor objects (e.g., furniture), and the RefCOCO annotations mainly describe humans, thereby intensifying category distribution mismatches; \textit{iii)} there exist significant scale differences, with ScanNet scenes typically around 5 meters and COCO-3D scenes ranging from 1 meter to several hundred meters; even after filtering out scenes larger than 50 meters, these scale discrepancies continue to affect transfer performance.

Nevertheless, once both COCO-3D and ScanNet are jointly used during training and the model is subsequently tested on ScanNet, the resulting performance exceeds the baseline by a notable margin (exceeding 6\%). This demonstrates that, despite marked differences in environment types, class distributions, and scales, synthetic data can effectively improve model generalization and overall performance by enriching the diversity of training data. Additionally, training and testing on COCO-3D alone still yield a discernible level of referring segmentation capability, although this level is below that obtained on real data. A reasonable explanation is that RefCOCO annotations are originally designed for 2D images, and directly applying them to 3D descriptions introduces certain inconsistencies that inevitably affect performance.

\subsection{3D LLM Tasks}
LLM development has advanced rapidly due to the abundance of 2D images, resulting in effective applications. Extending these models to 3D tasks is a natural next step, although current 3D datasets remain limited in scale and diversity. We evaluate the effect of our synthetic data on 3D tasks by pre-training the generalist model of LL3DA \cite{chen2023ll3da} on COCO-3D and fine-tuning it on ScanRefer, Nr3D, and ScanQA for 3D question answering and dense captioning. As shown in \cref{tab:ll3da_gen}, pre-training on COCO-3D leads to consistent improvements in multiple metrics compared with training solely on real-world data. In the ScanRefer task, metrics such as C@0.5 and B-4@0.5 increase, indicating improved captioning quality. A similar trend is observed for Nr3D, suggesting that the diverse scene and semantic distributions in COCO-3D enhance the model's generalization.  Additionally, \cref{tab:ll3da_qa} compares the performance of different 3D LLM models on the 3D question answering task when fine-tuned only on ScanQA. The results indicate that, compared with using a 7B-parameter model (e.g., Vicuna-7B \cite{zheng2024judging}), fine-tuning the 1.3B-parameter LL3DA \cite{chen2023ll3da} model can yield similar or even better results, with the BLEU-4 score reaching 15.8\%.

\section{Conclusion}

In this work, we present a novel approach to bridging the data gap for spatial intelligence by generating high-quality spatial data from large-scale annotated 2D image datasets. Our method leverages existing 2D annotations to construct diverse, realistic, in-the-wild 3D scenes, enabling the creation of new, large-scale spatial datasets such as \textit{COCO-3D} and \textit{Objects365-v2-3D}. These datasets significantly expand the coverage of 3D scene data, offering a robust foundation for spatial intelligence and embodied AI. Our results demonstrate the potential of scalable 2D-to-3D lifting as a cost-effective solution for advancing spatial perception and reasoning tasks.

\vspace{+1.0mm}

\noindent
\textbf{Limitations:} While our generated spatial datasets are poised to advance spatial intelligence, our current focus has remained on spatial perception and reasoning. We have not explored interactive 3D environments, which are crucial for embodied AI and vision-language-action (VLA) models, because such tasks often require specialized robotics hardware. Nevertheless, we believe our datasets offer a solid starting point for future research on spatial interaction and VLA systems.

\clearpage
{
    \bibliographystyle{ieeenat_fullname}
    \bibliography{main}

\begin{thebibliography}{150}
\providecommand{\natexlab}[1]{#1}
\providecommand{\url}[1]{\texttt{#1}}
\expandafter\ifx\csname urlstyle\endcsname\relax
  \providecommand{\doi}[1]{doi: #1}\else
  \providecommand{\doi}{doi: \begingroup \urlstyle{rm}\Url}\fi

\bibitem[Aan{\ae}s et~al.(2016)Aan{\ae}s, Jensen, Vogiatzis, Tola, and Dahl]{aanaes2016large}
Henrik Aan{\ae}s, Rasmus~Ramsb{\o}l Jensen, George Vogiatzis, Engin Tola, and Anders~Bjorholm Dahl.
\newblock Large-scale data for multiple-view stereopsis.
\newblock \emph{International Journal of Computer Vision}, pages 1--16, 2016.

\bibitem[Achlioptas et~al.(2020{\natexlab{a}})Achlioptas, Abdelreheem, Xia, Elhoseiny, and Guibas]{achlioptas2020referit3d}
Panos Achlioptas, Ahmed Abdelreheem, Fei Xia, Mohamed Elhoseiny, and Leonidas Guibas.
\newblock Referit3d: Neural listeners for fine-grained 3d object identification in real-world scenes.
\newblock In \emph{Computer Vision--ECCV 2020: 16th European Conference, Glasgow, UK, August 23--28, 2020, Proceedings, Part I 16}, pages 422--440. Springer, 2020{\natexlab{a}}.

\bibitem[Achlioptas et~al.(2020{\natexlab{b}})Achlioptas, Abdelreheem, Xia, Elhoseiny, and Guibas]{achlioptas2020referit_3d}
Panos Achlioptas, Ahmed Abdelreheem, Fei Xia, Mohamed Elhoseiny, and Leonidas~J. Guibas.
\newblock {ReferIt3D}: Neural listeners for fine-grained 3d object identification in real-world scenes.
\newblock In \emph{16th European Conference on Computer Vision (ECCV)}, 2020{\natexlab{b}}.

\bibitem[Albardaner et~al.(2024)Albardaner, Miguel, Garc{\'\i}a, and Dalmau]{albardaner2024sim}
Jaume Albardaner, Alberto~San Miguel, N{\'e}stor Garc{\'\i}a, and Mag{\'\i} Dalmau.
\newblock Sim-to-real gap in rl: Use case with tiago and isaac sim/gym.
\newblock In \emph{European Robotics Forum}, pages 344--348. Springer, 2024.

\bibitem[Amaduzzi et~al.(2024)Amaduzzi, Zama~Ramirez, Lisanti, Salti, and Di~Stefano]{NeurIPS24}
Andrea Amaduzzi, Pierluigi Zama~Ramirez, Giuseppe Lisanti, Samuele Salti, and Luigi Di~Stefano.
\newblock {LLaNA}: Large language and {NeRF} assistant.
\newblock In \emph{Advances in Neural Information Processing Systems (NeurIPS)}, 2024.

\bibitem[Andersen et~al.(2012)Andersen, Jensen, Lisouski, Mortensen, Hansen, Gregersen, and Ahrendt]{andersen2012kinect}
Michael~Riis Andersen, Thomas Jensen, Pavel Lisouski, Anders~Krogh Mortensen, Mikkel~Kragh Hansen, Torben Gregersen, and PJAU Ahrendt.
\newblock Kinect depth sensor evaluation for computer vision applications.
\newblock \emph{Aarhus University}, pages 1--37, 2012.

\bibitem[Armeni et~al.(2016)Armeni, Sener, Zamir, Jiang, Brilakis, Fischer, and Savarese]{armeni20163d}
Iro Armeni, Ozan Sener, Amir~R Zamir, Helen Jiang, Ioannis Brilakis, Martin Fischer, and Silvio Savarese.
\newblock 3d semantic parsing of large-scale indoor spaces.
\newblock In \emph{Proceedings of the IEEE conference on computer vision and pattern recognition}, pages 1534--1543, 2016.

\bibitem[Azuma et~al.(2022{\natexlab{a}})Azuma, Miyanishi, Kurita, and Kawanabe]{azuma2022scanqa}
Daichi Azuma, Taiki Miyanishi, Shuhei Kurita, and Motoaki Kawanabe.
\newblock Scanqa: 3d question answering for spatial scene understanding.
\newblock In \emph{proceedings of the IEEE/CVF conference on computer vision and pattern recognition}, pages 19129--19139, 2022{\natexlab{a}}.

\bibitem[Azuma et~al.(2022{\natexlab{b}})Azuma, Miyanishi, Kurita, and Kawanabe]{azuma_2022_CVPR}
Daichi Azuma, Taiki Miyanishi, Shuhei Kurita, and Motoaki Kawanabe.
\newblock Scanqa: 3d question answering for spatial scene understanding.
\newblock In \emph{Proceedings of the IEEE/CVF Conference on Computer Vision and Pattern Recognition (CVPR)}, 2022{\natexlab{b}}.

\bibitem[Bahmani et~al.(2024)Bahmani, Liu, Yifan, Skorokhodov, Rong, Liu, Liu, Park, Tulyakov, Wetzstein, et~al.]{bahmani2024tc4d}
Sherwin Bahmani, Xian Liu, Wang Yifan, Ivan Skorokhodov, Victor Rong, Ziwei Liu, Xihui Liu, Jeong~Joon Park, Sergey Tulyakov, Gordon Wetzstein, et~al.
\newblock Tc4d: Trajectory-conditioned text-to-4d generation.
\newblock In \emph{European Conference on Computer Vision}, pages 53--72. Springer, 2024.

\bibitem[Baruch et~al.(2021)Baruch, Chen, Dehghan, Dimry, Feigin, Fu, Gebauer, Joffe, Kurz, Schwartz, et~al.]{baruch2021arkitscenes}
Gilad Baruch, Zhuoyuan Chen, Afshin Dehghan, Tal Dimry, Yuri Feigin, Peter Fu, Thomas Gebauer, Brandon Joffe, Daniel Kurz, Arik Schwartz, et~al.
\newblock Arkitscenes: A diverse real-world dataset for 3d indoor scene understanding using mobile rgb-d data.
\newblock \emph{arXiv preprint arXiv:2111.08897}, 2021.

\bibitem[Bhat et~al.(2021)Bhat, Alhashim, and Wonka]{bhat2021adabins}
Shariq~Farooq Bhat, Ibraheem Alhashim, and Peter Wonka.
\newblock Adabins: Depth estimation using adaptive bins.
\newblock In \emph{Proceedings of the IEEE/CVF conference on computer vision and pattern recognition}, pages 4009--4018, 2021.

\bibitem[Bhat et~al.(2023)Bhat, Birkl, Wofk, Wonka, and M{\"u}ller]{bhat2023zoedepth}
Shariq~Farooq Bhat, Reiner Birkl, Diana Wofk, Peter Wonka, and Matthias M{\"u}ller.
\newblock Zoedepth: Zero-shot transfer by combining relative and metric depth.
\newblock \emph{arXiv preprint arXiv:2302.12288}, 2023.

\bibitem[Bian et~al.(2024)Bian, Kong, Xie, Pan, Qiao, and Liu]{bian2024dynamiccity}
Hengwei Bian, Lingdong Kong, Haozhe Xie, Liang Pan, Yu Qiao, and Ziwei Liu.
\newblock Dynamiccity: Large-scale lidar generation from dynamic scenes.
\newblock \emph{arXiv preprint arXiv:2410.18084}, 2024.

\bibitem[Black et~al.(2024)Black, Brown, Driess, Esmail, Equi, Finn, Fusai, Groom, Hausman, Ichter, et~al.]{black2024pi_0}
Kevin Black, Noah Brown, Danny Driess, Adnan Esmail, Michael Equi, Chelsea Finn, Niccolo Fusai, Lachy Groom, Karol Hausman, Brian Ichter, et~al.
\newblock $pi\_0$: A vision-language-action flow model for general robot control.
\newblock \emph{arXiv preprint arXiv:2410.24164}, 2024.

\bibitem[Brazil et~al.(2023)Brazil, Kumar, Straub, Ravi, Johnson, and Gkioxari]{brazil2023omni3d}
Garrick Brazil, Abhinav Kumar, Julian Straub, Nikhila Ravi, Justin Johnson, and Georgia Gkioxari.
\newblock Omni3d: A large benchmark and model for 3d object detection in the wild.
\newblock In \emph{Proceedings of the IEEE/CVF conference on computer vision and pattern recognition}, pages 13154--13164, 2023.

\bibitem[Brohan et~al.(2023)Brohan, Brown, Carbajal, Chebotar, Chen, Choromanski, Ding, Driess, Dubey, Finn, Florence, Fu, Arenas, Gopalakrishnan, Han, Hausman, Herzog, Hsu, Ichter, Irpan, Joshi, Julian, Kalashnikov, Kuang, Leal, Lee, Lee, Levine, Lu, Michalewski, Mordatch, Pertsch, Rao, Reymann, Ryoo, Salazar, Sanketi, Sermanet, Singh, Singh, Soricut, Tran, Vanhoucke, Vuong, Wahid, Welker, Wohlhart, Wu, Xia, Xiao, Xu, Xu, Yu, and Zitkovich]{rt22023arxiv}
Anthony Brohan, Noah Brown, Justice Carbajal, Yevgen Chebotar, Xi Chen, Krzysztof Choromanski, Tianli Ding, Danny Driess, Avinava Dubey, Chelsea Finn, Pete Florence, Chuyuan Fu, Montse~Gonzalez Arenas, Keerthana Gopalakrishnan, Kehang Han, Karol Hausman, Alex Herzog, Jasmine Hsu, Brian Ichter, Alex Irpan, Nikhil Joshi, Ryan Julian, Dmitry Kalashnikov, Yuheng Kuang, Isabel Leal, Lisa Lee, Tsang-Wei~Edward Lee, Sergey Levine, Yao Lu, Henryk Michalewski, Igor Mordatch, Karl Pertsch, Kanishka Rao, Krista Reymann, Michael Ryoo, Grecia Salazar, Pannag Sanketi, Pierre Sermanet, Jaspiar Singh, Anikait Singh, Radu Soricut, Huong Tran, Vincent Vanhoucke, Quan Vuong, Ayzaan Wahid, Stefan Welker, Paul Wohlhart, Jialin Wu, Fei Xia, Ted Xiao, Peng Xu, Sichun Xu, Tianhe Yu, and Brianna Zitkovich.
\newblock Rt-2: Vision-language-action models transfer web knowledge to robotic control.
\newblock In \emph{arXiv preprint arXiv:2307.15818}, 2023.

\bibitem[Cai et~al.(2024)Cai, Ponomarenko, Yuan, Li, Yang, Dong, and Zhao]{cai2024spatialbot}
Wenxiao Cai, Yaroslav Ponomarenko, Jianhao Yuan, Xiaoqi Li, Wankou Yang, Hao Dong, and Bo Zhao.
\newblock Spatialbot: Precise spatial understanding with vision language models.
\newblock \emph{arXiv preprint arXiv:2406.13642}, 2024.

\bibitem[Cao et~al.(2024)Cao, Pan, Han, Wong, and Liu]{cao2024avatargo}
Yukang Cao, Liang Pan, Kai Han, Kwan-Yee~K Wong, and Ziwei Liu.
\newblock Avatargo: Zero-shot 4d human-object interaction generation and animation.
\newblock \emph{arXiv preprint arXiv:2410.07164}, 2024.

\bibitem[Chang et~al.(2017)Chang, Dai, Funkhouser, Halber, Niessner, Savva, Song, Zeng, and Zhang]{Matterport3D}
Angel Chang, Angela Dai, Thomas Funkhouser, Maciej Halber, Matthias Niessner, Manolis Savva, Shuran Song, Andy Zeng, and Yinda Zhang.
\newblock Matterport3d: Learning from rgb-d data in indoor environments.
\newblock \emph{International Conference on 3D Vision (3DV)}, 2017.

\bibitem[Chang et~al.(2015)Chang, Funkhouser, Guibas, Hanrahan, Huang, Li, Savarese, Savva, Song, Su, et~al.]{chang2015shapenet}
Angel~X Chang, Thomas Funkhouser, Leonidas Guibas, Pat Hanrahan, Qixing Huang, Zimo Li, Silvio Savarese, Manolis Savva, Shuran Song, Hao Su, et~al.
\newblock Shapenet: An information-rich 3d model repository.
\newblock \emph{arXiv preprint arXiv:1512.03012}, 2015.

\bibitem[Chen et~al.(2024{\natexlab{a}})Chen, Xu, Kirmani, Ichter, Sadigh, Guibas, and Xia]{Chen_2024_CVPR}
Boyuan Chen, Zhuo Xu, Sean Kirmani, Brain Ichter, Dorsa Sadigh, Leonidas Guibas, and Fei Xia.
\newblock Spatialvlm: Endowing vision-language models with spatial reasoning capabilities.
\newblock In \emph{Proceedings of the IEEE/CVF Conference on Computer Vision and Pattern Recognition (CVPR)}, pages 14455--14465, 2024{\natexlab{a}}.

\bibitem[Chen et~al.(2024{\natexlab{b}})Chen, Xu, Kirmani, Ichter, Sadigh, Guibas, and Xia]{chen2024spatialvlm}
Boyuan Chen, Zhuo Xu, Sean Kirmani, Brain Ichter, Dorsa Sadigh, Leonidas Guibas, and Fei Xia.
\newblock Spatialvlm: Endowing vision-language models with spatial reasoning capabilities.
\newblock In \emph{Proceedings of the IEEE/CVF Conference on Computer Vision and Pattern Recognition}, pages 14455--14465, 2024{\natexlab{b}}.

\bibitem[Chen et~al.(2020)Chen, Chang, and Nie{\ss}ner]{chen2020scanrefer}
Dave~Zhenyu Chen, Angel~X Chang, and Matthias Nie{\ss}ner.
\newblock Scanrefer: 3d object localization in rgb-d scans using natural language.
\newblock \emph{16th European Conference on Computer Vision (ECCV)}, 2020.

\bibitem[Chen et~al.(2023)Chen, Chen, Zhang, Li, Yu, Fei, Zhu, Fan, and Chen]{chen2023ll3da}
Sijin Chen, Xin Chen, Chi Zhang, Mingsheng Li, Gang Yu, Hao Fei, Hongyuan Zhu, Jiayuan Fan, and Tao Chen.
\newblock Ll3da: Visual interactive instruction tuning for omni-3d understanding, reasoning, and planning, 2023.

\bibitem[Chen et~al.(2024{\natexlab{c}})Chen, Chen, Zhang, Li, Yu, Fei, Zhu, Fan, and Chen]{chen2024ll3da}
Sijin Chen, Xin Chen, Chi Zhang, Mingsheng Li, Gang Yu, Hao Fei, Hongyuan Zhu, Jiayuan Fan, and Tao Chen.
\newblock Ll3da: Visual interactive instruction tuning for omni-3d understanding reasoning and planning.
\newblock In \emph{Proceedings of the IEEE/CVF Conference on Computer Vision and Pattern Recognition}, pages 26428--26438, 2024{\natexlab{c}}.

\bibitem[Chen et~al.(2025)Chen, Chen, Pang, Zeng, Cheng, Fu, Yin, Wang, Yu, Yu, et~al.]{chen2025meshxl}
Sijin Chen, Xin Chen, Anqi Pang, Xianfang Zeng, Wei Cheng, Yijun Fu, Fukun Yin, Billzb Wang, Jingyi Yu, Gang Yu, et~al.
\newblock Meshxl: Neural coordinate field for generative 3d foundation models.
\newblock \emph{Advances in Neural Information Processing Systems}, 37:\penalty0 97141--97166, 2025.

\bibitem[Chen et~al.(2021)Chen, Gholami, Nie{\ss}ner, and Chang]{chen2021scan2cap}
Zhenyu Chen, Ali Gholami, Matthias Nie{\ss}ner, and Angel~X Chang.
\newblock Scan2cap: Context-aware dense captioning in rgb-d scans.
\newblock In \emph{Proceedings of the IEEE/CVF conference on computer vision and pattern recognition}, pages 3193--3203, 2021.

\bibitem[Cheng et~al.(2024)Cheng, Yin, Fu, Guo, Yang, Kautz, Wang, and Liu]{cheng2024spatialrgpt}
An-Chieh Cheng, Hongxu Yin, Yang Fu, Qiushan Guo, Ruihan Yang, Jan Kautz, Xiaolong Wang, and Sifei Liu.
\newblock Spatialrgpt: Grounded spatial reasoning in vision-language models.
\newblock In \emph{NeurIPS}, 2024.

\bibitem[Chung et~al.(2023{\natexlab{a}})Chung, Lee, Nam, Lee, and Lee]{chung2023lucid}
Jaeyoung Chung, Suyoung Lee, Hyeongjin Nam, Jaerin Lee, and Kyoung~Mu Lee.
\newblock Luciddreamer: Domain-free generation of 3d gaussian splatting scenes.
\newblock \emph{arXiv preprint arXiv:2311.13384}, 2023{\natexlab{a}}.

\bibitem[Chung et~al.(2023{\natexlab{b}})Chung, Lee, Nam, Lee, and Lee]{chung2023luciddreamer}
Jaeyoung Chung, Suyoung Lee, Hyeongjin Nam, Jaerin Lee, and Kyoung~Mu Lee.
\newblock Luciddreamer: Domain-free generation of 3d gaussian splatting scenes.
\newblock \emph{arXiv preprint arXiv:2311.13384}, 2023{\natexlab{b}}.

\bibitem[Collins et~al.(2022)Collins, Goel, Deng, Luthra, Xu, Gundogdu, Zhang, Vicente, Dideriksen, Arora, et~al.]{collins2022abo}
Jasmine Collins, Shubham Goel, Kenan Deng, Achleshwar Luthra, Leon Xu, Erhan Gundogdu, Xi Zhang, Tomas F~Yago Vicente, Thomas Dideriksen, Himanshu Arora, et~al.
\newblock Abo: Dataset and benchmarks for real-world 3d object understanding.
\newblock In \emph{Proceedings of the IEEE/CVF conference on computer vision and pattern recognition}, pages 21126--21136, 2022.

\bibitem[Contributors(2023)]{pointcept2023}
Pointcept Contributors.
\newblock Pointcept: A codebase for point cloud perception research.
\newblock \url{https://github.com/Pointcept/Pointcept}, 2023.

\bibitem[Contributors(2022)]{spconv2022}
Spconv Contributors.
\newblock Spconv: Spatially sparse convolution library.
\newblock \url{https://github.com/traveller59/spconv}, 2022.

\bibitem[Dai et~al.(2017)Dai, Chang, Savva, Halber, Funkhouser, and Nie{\ss}ner]{dai2017scannet}
Angela Dai, Angel~X Chang, Manolis Savva, Maciej Halber, Thomas Funkhouser, and Matthias Nie{\ss}ner.
\newblock Scannet: Richly-annotated 3d reconstructions of indoor scenes.
\newblock In \emph{Proceedings of the IEEE conference on computer vision and pattern recognition}, pages 5828--5839, 2017.

\bibitem[Devlin(2018)]{devlin2018bert}
Jacob Devlin.
\newblock Bert: Pre-training of deep bidirectional transformers for language understanding.
\newblock \emph{arXiv preprint arXiv:1810.04805}, 2018.

\bibitem[Downs et~al.(2022)Downs, Francis, Koenig, Kinman, Hickman, Reymann, McHugh, and Vanhoucke]{downs2022google}
Laura Downs, Anthony Francis, Nate Koenig, Brandon Kinman, Ryan Hickman, Krista Reymann, Thomas~B McHugh, and Vincent Vanhoucke.
\newblock Google scanned objects: A high-quality dataset of 3d scanned household items.
\newblock In \emph{2022 International Conference on Robotics and Automation (ICRA)}, pages 2553--2560. IEEE, 2022.

\bibitem[Duan et~al.(2023)Duan, Long, Wang, Zhang, Willcocks, and Shao]{duan2023dynamic}
Haoran Duan, Yang Long, Shidong Wang, Haofeng Zhang, Chris~G Willcocks, and Ling Shao.
\newblock Dynamic unary convolution in transformers.
\newblock \emph{IEEE Transactions on Pattern Analysis and Machine Intelligence}, 45\penalty0 (11):\penalty0 12747--12759, 2023.

\bibitem[Duan et~al.(2022)Duan, Yu, Tan, Zhu, and Tan]{duan2022survey}
Jiafei Duan, Samson Yu, Hui~Li Tan, Hongyuan Zhu, and Cheston Tan.
\newblock A survey of embodied ai: From simulators to research tasks.
\newblock \emph{IEEE Transactions on Emerging Topics in Computational Intelligence}, 6\penalty0 (2):\penalty0 230--244, 2022.

\bibitem[Eigen et~al.(2014)Eigen, Puhrsch, and Fergus]{eigen2014depth}
David Eigen, Christian Puhrsch, and Rob Fergus.
\newblock Depth map prediction from a single image using a multi-scale deep network.
\newblock \emph{Advances in neural information processing systems}, 27, 2014.

\bibitem[Fan et~al.(2024)Fan, Li, Ding, Zhou, and Qian]{fan2024integrating}
Chao Fan, Zihan Li, Weike Ding, Huiming Zhou, and Kun Qian.
\newblock Integrating artificial intelligence with slam technology for robotic navigation and localization in unknown environments.
\newblock \emph{International Journal of Robotics and Automation}, 29\penalty0 (4):\penalty0 215--230, 2024.

\bibitem[Fu et~al.(2021)Fu, Jia, Gao, Gong, Zhao, Maybank, and Tao]{fu20213d}
Huan Fu, Rongfei Jia, Lin Gao, Mingming Gong, Binqiang Zhao, Steve Maybank, and Dacheng Tao.
\newblock 3d-future: 3d furniture shape with texture.
\newblock \emph{International Journal of Computer Vision}, 129:\penalty0 3313--3337, 2021.

\bibitem[Fu et~al.(2024)Fu, Liu, Chen, Nie, and Xiong]{fu2024scene}
Rao Fu, Jingyu Liu, Xilun Chen, Yixin Nie, and Wenhan Xiong.
\newblock Scene-llm: Extending language model for 3d visual understanding and reasoning.
\newblock \emph{arXiv preprint arXiv:2403.11401}, 2024.

\bibitem[Geiger et~al.(2013)Geiger, Lenz, Stiller, and Urtasun]{geiger2013vision}
Andreas Geiger, Philip Lenz, Christoph Stiller, and Raquel Urtasun.
\newblock Vision meets robotics: The kitti dataset.
\newblock \emph{The International Journal of Robotics Research}, 32\penalty0 (11):\penalty0 1231--1237, 2013.

\bibitem[Gulino et~al.(2023)Gulino, Fu, Luo, Tucker, Bronstein, Lu, Harb, Pan, Wang, Chen, et~al.]{gulino2023waymax}
Cole Gulino, Justin Fu, Wenjie Luo, George Tucker, Eli Bronstein, Yiren Lu, Jean Harb, Xinlei Pan, Yan Wang, Xiangyu Chen, et~al.
\newblock Waymax: An accelerated, data-driven simulator for large-scale autonomous driving research.
\newblock \emph{Advances in Neural Information Processing Systems}, 36:\penalty0 7730--7742, 2023.

\bibitem[Guo et~al.(2023{\natexlab{a}})Guo, Tang, Zhang, Wang, Wang, Zhao, and Li]{guo2023viewrefer}
Ziyu Guo, Yiwen Tang, Renrui Zhang, Dong Wang, Zhigang Wang, Bin Zhao, and Xuelong Li.
\newblock Viewrefer: Grasp the multi-view knowledge for 3d visual grounding with gpt and prototype guidance.
\newblock \emph{arXiv preprint arXiv:2303.16894}, 2023{\natexlab{a}}.

\bibitem[Guo et~al.(2023{\natexlab{b}})Guo, Zhang, Zhu, Tang, Ma, Han, Chen, Gao, Li, Li, et~al.]{guo2023point}
Ziyu Guo, Renrui Zhang, Xiangyang Zhu, Yiwen Tang, Xianzheng Ma, Jiaming Han, Kexin Chen, Peng Gao, Xianzhi Li, Hongsheng Li, et~al.
\newblock Point-bind \& point-llm: Aligning point cloud with multi-modality for 3d understanding, generation, and instruction following.
\newblock \emph{arXiv preprint arXiv:2309.00615}, 2023{\natexlab{b}}.

\bibitem[Haghighi et~al.(2024)Haghighi, Samadi, Dianati, Donzella, and Debattista]{haghighi2024taming}
Hamed Haghighi, Amir Samadi, Mehrdad Dianati, Valentina Donzella, and Kurt Debattista.
\newblock Taming transformers for realistic lidar point cloud generation.
\newblock \emph{arXiv preprint arXiv:2404.05505}, 2024.

\bibitem[He et~al.(2024)He, Chen, Peng, Huang, Li, Huang, Yuan, Ouyang, and He]{he2024gvgen}
Xianglong He, Junyi Chen, Sida Peng, Di Huang, Yangguang Li, Xiaoshui Huang, Chun Yuan, Wanli Ouyang, and Tong He.
\newblock Gvgen: Text-to-3d generation with volumetric representation.
\newblock In \emph{European Conference on Computer Vision}, pages 463--479. Springer, 2024.

\bibitem[Hong et~al.(2024)Hong, Tang, Cao, Shi, Wu, Chen, Yang, Wang, Pan, Lin, et~al.]{hong20243dtopia}
Fangzhou Hong, Jiaxiang Tang, Ziang Cao, Min Shi, Tong Wu, Zhaoxi Chen, Shuai Yang, Tengfei Wang, Liang Pan, Dahua Lin, et~al.
\newblock 3dtopia: Large text-to-3d generation model with hybrid diffusion priors.
\newblock \emph{arXiv preprint arXiv:2403.02234}, 2024.

\bibitem[Hong et~al.(2023{\natexlab{a}})Hong, Zhang, Gu, Bi, Zhou, Liu, Liu, Sunkavalli, Bui, and Tan]{hong2023lrm}
Yicong Hong, Kai Zhang, Jiuxiang Gu, Sai Bi, Yang Zhou, Difan Liu, Feng Liu, Kalyan Sunkavalli, Trung Bui, and Hao Tan.
\newblock Lrm: Large reconstruction model for single image to 3d.
\newblock \emph{arXiv preprint arXiv:2311.04400}, 2023{\natexlab{a}}.

\bibitem[Hong et~al.(2023{\natexlab{b}})Hong, Zhen, Chen, Zheng, Du, Chen, and Gan]{3dllm}
Yining Hong, Haoyu Zhen, Peihao Chen, Shuhong Zheng, Yilun Du, Zhenfang Chen, and Chuang Gan.
\newblock 3d-llm: Injecting the 3d world into large language models.
\newblock \emph{NeurIPS}, 2023{\natexlab{b}}.

\bibitem[Hu et~al.(2024)Hu, Yin, Zhang, Cai, Long, Chen, Wang, Yu, Shen, and Shen]{hu2024metric3d}
Mu Hu, Wei Yin, Chi Zhang, Zhipeng Cai, Xiaoxiao Long, Hao Chen, Kaixuan Wang, Gang Yu, Chunhua Shen, and Shaojie Shen.
\newblock Metric3d v2: A versatile monocular geometric foundation model for zero-shot metric depth and surface normal estimation.
\newblock \emph{arXiv preprint arXiv:2404.15506}, 2024.

\bibitem[Hua et~al.(2025)Hua, Liu, Su, Miao, Ouyang, Wang, Hu, Wen, Zhai, Long, et~al.]{hua2025attention}
Litao Hua, Fan Liu, Jie Su, Xingyu Miao, Zizhou Ouyang, Zeyu Wang, Runze Hu, Zhenyu Wen, Bing Zhai, Yang Long, et~al.
\newblock Attention in diffusion model: A survey.
\newblock \emph{arXiv preprint arXiv:2504.03738}, 2025.

\bibitem[Huang et~al.(2023{\natexlab{a}})Huang, Wang, Huang, Liu, Cheng, Zhao, Jin, and Zhao]{huang2023chat}
Haifeng Huang, Zehan Wang, Rongjie Huang, Luping Liu, Xize Cheng, Yang Zhao, Tao Jin, and Zhou Zhao.
\newblock Chat-3d v2: Bridging 3d scene and large language models with object identifiers.
\newblock \emph{arXiv preprint arXiv:2312.08168}, 2023{\natexlab{a}}.

\bibitem[Huang et~al.(2024{\natexlab{a}})Huang, Chen, Wang, Huang, Xu, Wang, Liu, Cheng, Zhao, Pang, et~al.]{huang2024chat}
Haifeng Huang, Yilun Chen, Zehan Wang, Rongjie Huang, Runsen Xu, Tai Wang, Luping Liu, Xize Cheng, Yang Zhao, Jiangmiao Pang, et~al.
\newblock Chat-scene: Bridging 3d scene and large language models with object identifiers.
\newblock \emph{Proceedings of the Advances in Neural Information Processing Systems, Vancouver, BC, Canada}, 2024{\natexlab{a}}.

\bibitem[Huang et~al.(2023{\natexlab{b}})Huang, Yong, Ma, Linghu, Li, Wang, Li, Zhu, Jia, and Huang]{huang2023embodied}
Jiangyong Huang, Silong Yong, Xiaojian Ma, Xiongkun Linghu, Puhao Li, Yan Wang, Qing Li, Song-Chun Zhu, Baoxiong Jia, and Siyuan Huang.
\newblock An embodied generalist agent in 3d world.
\newblock \emph{arXiv preprint arXiv:2311.12871}, 2023{\natexlab{b}}.

\bibitem[Huang et~al.(2021)Huang, Lee, Chen, and Liu]{huang2021text}
Pin-Hao Huang, Han-Hung Lee, Hwann-Tzong Chen, and Tyng-Luh Liu.
\newblock Text-guided graph neural networks for referring 3d instance segmentation.
\newblock In \emph{Proceedings of the AAAI Conference on Artificial Intelligence}, pages 1610--1618, 2021.

\bibitem[Huang et~al.(2024{\natexlab{b}})Huang, Wu, Chen, Zhao, Zhu, and Lasenby]{huang2024openins3d}
Zhening Huang, Xiaoyang Wu, Xi Chen, Hengshuang Zhao, Lei Zhu, and Joan Lasenby.
\newblock Openins3d: Snap and lookup for 3d open-vocabulary instance segmentation.
\newblock In \emph{European Conference on Computer Vision}, pages 169--185. Springer, 2024{\natexlab{b}}.

\bibitem[Jin et~al.(2023)Jin, Zhang, Hold-Geoffroy, Wang, Matzen, Sticha, and Fouhey]{jin2023perspective}
Linyi Jin, Jianming Zhang, Yannick Hold-Geoffroy, Oliver Wang, Kevin Matzen, Matthew Sticha, and David~F. Fouhey.
\newblock Perspective fields for single image camera calibration.
\newblock In \emph{CVPR}, 2023.

\bibitem[Ke et~al.(2024)Ke, Obukhov, Huang, Metzger, Daudt, and Schindler]{ke2023repurposing}
Bingxin Ke, Anton Obukhov, Shengyu Huang, Nando Metzger, Rodrigo~Caye Daudt, and Konrad Schindler.
\newblock Repurposing diffusion-based image generators for monocular depth estimation.
\newblock In \emph{Proceedings of the IEEE/CVF Conference on Computer Vision and Pattern Recognition (CVPR)}, 2024.

\bibitem[Kim et~al.(2024)Kim, Pertsch, Karamcheti, Xiao, Balakrishna, Nair, Rafailov, Foster, Lam, Sanketi, Vuong, Kollar, Burchfiel, Tedrake, Sadigh, Levine, Liang, and Finn]{kim24openvla}
{Moo Jin} Kim, Karl Pertsch, Siddharth Karamcheti, Ted Xiao, Ashwin Balakrishna, Suraj Nair, Rafael Rafailov, Ethan Foster, Grace Lam, Pannag Sanketi, Quan Vuong, Thomas Kollar, Benjamin Burchfiel, Russ Tedrake, Dorsa Sadigh, Sergey Levine, Percy Liang, and Chelsea Finn.
\newblock Openvla: An open-source vision-language-action model.
\newblock \emph{arXiv preprint arXiv:2406.09246}, 2024.

\bibitem[Kirillov et~al.(2023)Kirillov, Mintun, Ravi, Mao, Rolland, Gustafson, Xiao, Whitehead, Berg, Lo, et~al.]{kirillov2023segment}
Alexander Kirillov, Eric Mintun, Nikhila Ravi, Hanzi Mao, Chloe Rolland, Laura Gustafson, Tete Xiao, Spencer Whitehead, Alexander~C Berg, Wan-Yen Lo, et~al.
\newblock Segment anything.
\newblock In \emph{Proceedings of the IEEE/CVF international conference on computer vision}, pages 4015--4026, 2023.

\bibitem[Knapitsch et~al.(2017)Knapitsch, Park, Zhou, and Koltun]{Knapitsch2017}
Arno Knapitsch, Jaesik Park, Qian-Yi Zhou, and Vladlen Koltun.
\newblock Tanks and temples: Benchmarking large-scale scene reconstruction.
\newblock \emph{ACM Transactions on Graphics}, 36\penalty0 (4), 2017.

\bibitem[Kuznetsova et~al.(2020)Kuznetsova, Rom, Alldrin, Uijlings, Krasin, Pont-Tuset, Kamali, Popov, Malloci, Kolesnikov, et~al.]{kuznetsova2020open}
Alina Kuznetsova, Hassan Rom, Neil Alldrin, Jasper Uijlings, Ivan Krasin, Jordi Pont-Tuset, Shahab Kamali, Stefan Popov, Matteo Malloci, Alexander Kolesnikov, et~al.
\newblock The open images dataset v4: Unified image classification, object detection, and visual relationship detection at scale.
\newblock \emph{International journal of computer vision}, 128\penalty0 (7):\penalty0 1956--1981, 2020.

\bibitem[Lai et~al.(2023)Lai, Yuan, Chu, Chen, Hu, and Jia]{lai2023mask}
Xin Lai, Yuhui Yuan, Ruihang Chu, Yukang Chen, Han Hu, and Jiaya Jia.
\newblock Mask-attention-free transformer for 3d instance segmentation.
\newblock In \emph{Proceedings of the IEEE/CVF International Conference on Computer Vision}, pages 3693--3703, 2023.

\bibitem[Lan et~al.(2024)Lan, Hong, Yang, Zhou, Meng, Dai, Pan, and Loy]{lan2024ln3diff}
Yushi Lan, Fangzhou Hong, Shuai Yang, Shangchen Zhou, Xuyi Meng, Bo Dai, Xingang Pan, and Chen~Change Loy.
\newblock Ln3diff: Scalable latent neural fields diffusion for speedy 3d generation.
\newblock In \emph{European Conference on Computer Vision}, pages 112--130. Springer, 2024.

\bibitem[Li et~al.(2024{\natexlab{a}})Li, Zhang, Guo, Zhang, Li, Zhang, Zhang, Zhang, Li, Liu, et~al.]{li2024llava}
Bo Li, Yuanhan Zhang, Dong Guo, Renrui Zhang, Feng Li, Hao Zhang, Kaichen Zhang, Peiyuan Zhang, Yanwei Li, Ziwei Liu, et~al.
\newblock Llava-onevision: Easy visual task transfer.
\newblock \emph{arXiv preprint arXiv:2408.03326}, 2024{\natexlab{a}}.

\bibitem[Li et~al.(2024{\natexlab{b}})Li, Shi, Zhang, Wu, Liao, Wang, Lee, and Zhou]{li2024dreamscene}
Haoran Li, Haolin Shi, Wenli Zhang, Wenjun Wu, Yong Liao, Lin Wang, Lik-hang Lee, and Peng~Yuan Zhou.
\newblock Dreamscene: 3d gaussian-based text-to-3d scene generation via formation pattern sampling.
\newblock In \emph{European Conference on Computer Vision}, pages 214--230. Springer, 2024{\natexlab{b}}.

\bibitem[Li et~al.(2023)Li, Li, Savarese, and Hoi]{li2023blip}
Junnan Li, Dongxu Li, Silvio Savarese, and Steven Hoi.
\newblock Blip-2: Bootstrapping language-image pre-training with frozen image encoders and large language models.
\newblock In \emph{International conference on machine learning}, pages 19730--19742. PMLR, 2023.

\bibitem[Li et~al.(2024{\natexlab{c}})Li, Wang, He, Li, Wang, Liu, Wang, Xu, Chen, Luo, et~al.]{li2024mvbench}
Kunchang Li, Yali Wang, Yinan He, Yizhuo Li, Yi Wang, Yi Liu, Zun Wang, Jilan Xu, Guo Chen, Ping Luo, et~al.
\newblock Mvbench: A comprehensive multi-modal video understanding benchmark.
\newblock In \emph{Proceedings of the IEEE/CVF Conference on Computer Vision and Pattern Recognition}, pages 22195--22206, 2024{\natexlab{c}}.

\bibitem[Li et~al.(2021)Li, Yu, Sang, Wang, Song, Liu, Yeh, Zhu, Gundavarapu, Shi, Bi, Xu, Yu, Sunkavalli, Hašan, Ramamoorthi, and Chandraker]{li2021openroomsendtoendopenframework}
Zhengqin Li, Ting-Wei Yu, Shen Sang, Sarah Wang, Meng Song, Yuhan Liu, Yu-Ying Yeh, Rui Zhu, Nitesh Gundavarapu, Jia Shi, Sai Bi, Zexiang Xu, Hong-Xing Yu, Kalyan Sunkavalli, Miloš Hašan, Ravi Ramamoorthi, and Manmohan Chandraker.
\newblock Openrooms: An end-to-end open framework for photorealistic indoor scene datasets, 2021.

\bibitem[Li et~al.(2024{\natexlab{d}})Li, Li, Cui, Pollefeys, and Oswald]{li2024sat2scene}
Zuoyue Li, Zhenqiang Li, Zhaopeng Cui, Marc Pollefeys, and Martin~R Oswald.
\newblock Sat2scene: 3d urban scene generation from satellite images with diffusion.
\newblock In \emph{Proceedings of the IEEE/CVF Conference on Computer Vision and Pattern Recognition}, pages 7141--7150, 2024{\natexlab{d}}.

\bibitem[Li et~al.(2024{\natexlab{e}})Li, Wang, Liu, and Jiang]{li2024binsformer}
Zhenyu Li, Xuyang Wang, Xianming Liu, and Junjun Jiang.
\newblock Binsformer: Revisiting adaptive bins for monocular depth estimation.
\newblock \emph{IEEE Transactions on Image Processing}, 2024{\natexlab{e}}.

\bibitem[Liang et~al.(2024)Liang, Su, Schulter, Garg, Zhao, Wu, and Chandraker]{liang2024aide}
Mingfu Liang, Jong-Chyi Su, Samuel Schulter, Sparsh Garg, Shiyu Zhao, Ying Wu, and Manmohan Chandraker.
\newblock Aide: An automatic data engine for object detection in autonomous driving.
\newblock In \emph{Proceedings of the IEEE/CVF Conference on Computer Vision and Pattern Recognition}, pages 14695--14706, 2024.

\bibitem[Lin et~al.(2025)Lin, Pan, Yang, Li, and Mu]{lin2025diffsplat}
Chenguo Lin, Panwang Pan, Bangbang Yang, Zeming Li, and Yadong Mu.
\newblock Diffsplat: Repurposing image diffusion models for scalable 3d gaussian splat generation.
\newblock In \emph{International Conference on Learning Representations (ICLR)}, 2025.

\bibitem[Lin et~al.(2014)Lin, Maire, Belongie, Hays, Perona, Ramanan, Doll{\'a}r, and Zitnick]{lin2014microsoft}
Tsung-Yi Lin, Michael Maire, Serge Belongie, James Hays, Pietro Perona, Deva Ramanan, Piotr Doll{\'a}r, and C~Lawrence Zitnick.
\newblock Microsoft coco: Common objects in context.
\newblock In \emph{Computer Vision--ECCV 2014: 13th European Conference, Zurich, Switzerland, September 6-12, 2014, Proceedings, Part V 13}, pages 740--755. Springer, 2014.

\bibitem[Linghu et~al.(2025)Linghu, Huang, Niu, Ma, Jia, and Huang]{linghu2025multi}
Xiongkun Linghu, Jiangyong Huang, Xuesong Niu, Xiaojian~Shawn Ma, Baoxiong Jia, and Siyuan Huang.
\newblock Multi-modal situated reasoning in 3d scenes.
\newblock \emph{Advances in Neural Information Processing Systems}, 37:\penalty0 140903--140936, 2025.

\bibitem[Liu et~al.(2023)Liu, Dong, Zhang, Luo, Gao, Huang, Gong, and Wang]{liu20233daxiesprompts}
Dingning Liu, Xiaomeng Dong, Renrui Zhang, Xu Luo, Peng Gao, Xiaoshui Huang, Yongshun Gong, and Zhihui Wang.
\newblock 3daxiesprompts: Unleashing the 3d spatial task capabilities of gpt-4v.
\newblock \emph{arXiv preprint arXiv:2312.09738}, 2023.

\bibitem[Liu et~al.(2022)Liu, Xu, Fu, Qian, Yu, Han, and Lu]{liu2022akb}
Liu Liu, Wenqiang Xu, Haoyuan Fu, Sucheng Qian, Qiaojun Yu, Yang Han, and Cewu Lu.
\newblock Akb-48: A real-world articulated object knowledge base.
\newblock In \emph{Proceedings of the IEEE/CVF Conference on Computer Vision and Pattern Recognition}, pages 14809--14818, 2022.

\bibitem[Lu et~al.(2023)Lu, Deng, Wang, He, and Zhang]{lu2023query}
Jiahao Lu, Jiacheng Deng, Chuxin Wang, Jianfeng He, and Tianzhu Zhang.
\newblock Query refinement transformer for 3d instance segmentation.
\newblock In \emph{Proceedings of the IEEE/CVF International Conference on Computer Vision}, pages 18516--18526, 2023.

\bibitem[Luo et~al.(2021)Luo, Cai, Zhou, Zhang, Zhao, Yi, Lu, Li, Zhang, and Liu]{Luo_2021_ICCV}
Zhipeng Luo, Zhongang Cai, Changqing Zhou, Gongjie Zhang, Haiyu Zhao, Shuai Yi, Shijian Lu, Hongsheng Li, Shanghang Zhang, and Ziwei Liu.
\newblock Unsupervised domain adaptive 3d detection with multi-level consistency.
\newblock In \emph{Proceedings of the IEEE/CVF International Conference on Computer Vision (ICCV)}, pages 8866--8875, 2021.

\bibitem[Luo et~al.(2023)Luo, Zhang, Zhou, Liu, Lu, and Pan]{luo2023transpillars}
Zhipeng Luo, Gongjie Zhang, Changqing Zhou, Tianrui Liu, Shijian Lu, and Liang Pan.
\newblock Transpillars: Coarse-to-fine aggregation for multi-frame 3d object detection.
\newblock In \emph{Proceedings of the IEEE/CVF Winter Conference on Applications of Computer Vision (WACV)}, pages 4230--4239, 2023.

\bibitem[Luo et~al.(2024)Luo, Zhang, Zhou, Wu, Tao, Lu, and Lu]{luo2024modeling}
Zhipeng Luo, Gongjie Zhang, Changqing Zhou, Zhonghua Wu, Qingyi Tao, Lewei Lu, and Shijian Lu.
\newblock Modeling continuous motion for 3d point cloud object tracking.
\newblock In \emph{Proceedings of the AAAI Conference on Artificial Intelligence (AAAI)}, pages 4026--4034, 2024.

\bibitem[Makoviychuk et~al.(2021)Makoviychuk, Wawrzyniak, Guo, Lu, Storey, Macklin, Hoeller, Rudin, Allshire, Handa, et~al.]{makoviychuk2021isaac}
Viktor Makoviychuk, Lukasz Wawrzyniak, Yunrong Guo, Michelle Lu, Kier Storey, Miles Macklin, David Hoeller, Nikita Rudin, Arthur Allshire, Ankur Handa, et~al.
\newblock Isaac gym: High performance gpu-based physics simulation for robot learning.
\newblock \emph{arXiv preprint arXiv:2108.10470}, 2021.

\bibitem[Man et~al.(2024)Man, Gui, and Wang]{man2024situation}
Yunze Man, Liang-Yan Gui, and Yu-Xiong Wang.
\newblock Situational awareness matters in 3d vision language reasoning.
\newblock In \emph{Proceedings of the IEEE/CVF Conference on Computer Vision and Pattern Recognition}, 2024.

\bibitem[Miao et~al.(2025)Miao, Duan, Long, and Han]{miao2025rethinking}
Xingyu Miao, Haoran Duan, Yang Long, and Jungong Han.
\newblock Rethinking score distilling sampling for 3d editing and generation.
\newblock \emph{arXiv preprint arXiv:2505.01888}, 2025.

\bibitem[Miceli-Barone et~al.(2023)Miceli-Barone, Lascarides, and Innes]{miceli2023dialogue}
Antonio~Valerio Miceli-Barone, Alex Lascarides, and Craig Innes.
\newblock Dialogue-based generation of self-driving simulation scenarios using large language models.
\newblock \emph{arXiv preprint arXiv:2310.17372}, 2023.

\bibitem[Mildenhall et~al.(2019)Mildenhall, Srinivasan, Ortiz-Cayon, Kalantari, Ramamoorthi, Ng, and Kar]{mildenhall2019llff}
Ben Mildenhall, Pratul~P. Srinivasan, Rodrigo Ortiz-Cayon, Nima~Khademi Kalantari, Ravi Ramamoorthi, Ren Ng, and Abhishek Kar.
\newblock Local light field fusion: Practical view synthesis with prescriptive sampling guidelines.
\newblock \emph{ACM Transactions on Graphics (TOG)}, 2019.

\bibitem[Nathan~Silberman and Fergus(2012)]{SilbermanECCV12}
Pushmeet~Kohli Nathan~Silberman, Derek~Hoiem and Rob Fergus.
\newblock Indoor segmentation and support inference from rgbd images.
\newblock In \emph{ECCV}, 2012.

\bibitem[Omotuyi et~al.(2024)Omotuyi, Hoeller, and Burnham]{nvidia2024}
Oyindamola Omotuyi, David Hoeller, and Ted Burnham.
\newblock Closing the sim-to-real gap: Training spot quadruped locomotion with nvidia isaac lab, 2024.
\newblock Accessed: 2025-02-16.

\bibitem[Peng et~al.(2018)Peng, Andrychowicz, Zaremba, and Abbeel]{peng2018sim}
Xue~Bin Peng, Marcin Andrychowicz, Wojciech Zaremba, and Pieter Abbeel.
\newblock Sim-to-real transfer of robotic control with dynamics randomization.
\newblock In \emph{2018 IEEE international conference on robotics and automation (ICRA)}, pages 3803--3810. IEEE, 2018.

\bibitem[Piccinelli et~al.(2024)Piccinelli, Yang, Sakaridis, Segu, Li, Van~Gool, and Yu]{piccinelli2024unidepth}
Luigi Piccinelli, Yung-Hsu Yang, Christos Sakaridis, Mattia Segu, Siyuan Li, Luc Van~Gool, and Fisher Yu.
\newblock Unidepth: Universal monocular metric depth estimation.
\newblock In \emph{Proceedings of the IEEE/CVF Conference on Computer Vision and Pattern Recognition}, pages 10106--10116, 2024.

\bibitem[Poole et~al.(2022)Poole, Jain, Barron, and Mildenhall]{poole2022dreamfusion}
Ben Poole, Ajay Jain, Jonathan~T Barron, and Ben Mildenhall.
\newblock Dreamfusion: Text-to-3d using 2d diffusion.
\newblock \emph{arXiv preprint arXiv:2209.14988}, 2022.

\bibitem[Qi et~al.(2024{\natexlab{a}})Qi, Dong, Zhang, Geng, Han, Ge, Yi, and Ma]{qi2024shapellm}
Zekun Qi, Runpei Dong, Shaochen Zhang, Haoran Geng, Chunrui Han, Zheng Ge, Li Yi, and Kaisheng Ma.
\newblock Shapellm: Universal 3d object understanding for embodied interaction.
\newblock \emph{arXiv preprint arXiv:2402.17766}, 2024{\natexlab{a}}.

\bibitem[Qi et~al.(2024{\natexlab{b}})Qi, Fang, Sun, Wu, Wu, Wang, Lin, and Zhao]{qi2024gpt4point}
Zhangyang Qi, Ye Fang, Zeyi Sun, Xiaoyang Wu, Tong Wu, Jiaqi Wang, Dahua Lin, and Hengshuang Zhao.
\newblock Gpt4point: A unified framework for point-language understanding and generation.
\newblock In \emph{Proceedings of the IEEE/CVF Conference on Computer Vision and Pattern Recognition}, pages 26417--26427, 2024{\natexlab{b}}.

\bibitem[Qi et~al.(2024{\natexlab{c}})Qi, Zhang, Fang, Wang, and Zhao]{GPT4Scene}
Zhangyang Qi, Zhixiong Zhang, Ye Fang, Jiaqi Wang, and Hengshuang Zhao.
\newblock Gpt4scene: Understand 3d scenes from videos with vision-language models.
\newblock \emph{arXiv preprint arXiv:2501.01428}, 2024{\natexlab{c}}.

\bibitem[Reizenstein et~al.(2021)Reizenstein, Shapovalov, Henzler, Sbordone, Labatut, and Novotny]{reizenstein2021commonobjects3dlargescale}
Jeremy Reizenstein, Roman Shapovalov, Philipp Henzler, Luca Sbordone, Patrick Labatut, and David Novotny.
\newblock Common objects in 3d: Large-scale learning and evaluation of real-life 3d category reconstruction, 2021.

\bibitem[Ren et~al.(2025)Ren, Xie, Mirzaei, Kreis, Liu, Torralba, Fidler, Kim, Ling, et~al.]{ren2025l4gm}
Jiawei Ren, Cheng Xie, Ashkan Mirzaei, Karsten Kreis, Ziwei Liu, Antonio Torralba, Sanja Fidler, Seung~Wook Kim, Huan Ling, et~al.
\newblock L4gm: Large 4d gaussian reconstruction model.
\newblock \emph{Advances in Neural Information Processing Systems}, 37:\penalty0 56828--56858, 2025.

\bibitem[Rozenberszki et~al.(2022)Rozenberszki, Litany, and Dai]{rozenberszki2022language}
David Rozenberszki, Or Litany, and Angela Dai.
\newblock Language-grounded indoor 3d semantic segmentation in the wild.
\newblock In \emph{European Conference on Computer Vision}, pages 125--141. Springer, 2022.

\bibitem[Sch\"ops et~al.(2017)Sch\"ops, Sch\"onberger, Galliani, Sattler, Schindler, Pollefeys, and Geiger]{schoeps2017cvpr}
Thomas Sch\"ops, Johannes~L. Sch\"onberger, Silvano Galliani, Torsten Sattler, Konrad Schindler, Marc Pollefeys, and Andreas Geiger.
\newblock A multi-view stereo benchmark with high-resolution images and multi-camera videos.
\newblock In \emph{Conference on Computer Vision and Pattern Recognition (CVPR)}, 2017.

\bibitem[Schult et~al.(2023)Schult, Engelmann, Hermans, Litany, Tang, and Leibe]{schult2023mask3d}
Jonas Schult, Francis Engelmann, Alexander Hermans, Or Litany, Siyu Tang, and Bastian Leibe.
\newblock Mask3d: Mask transformer for 3d semantic instance segmentation.
\newblock In \emph{2023 IEEE International Conference on Robotics and Automation (ICRA)}, pages 8216--8223. IEEE, 2023.

\bibitem[Shao et~al.(2019)Shao, Li, Zhang, Peng, Yu, Zhang, Li, and Sun]{shao2019objects365}
Shuai Shao, Zeming Li, Tianyuan Zhang, Chao Peng, Gang Yu, Xiangyu Zhang, Jing Li, and Jian Sun.
\newblock Objects365: A large-scale, high-quality dataset for object detection.
\newblock In \emph{Proceedings of the IEEE/CVF international conference on computer vision}, pages 8430--8439, 2019.

\bibitem[Song et~al.(2015)Song, Lichtenberg, and Xiao]{song2015sun}
Shuran Song, Samuel~P Lichtenberg, and Jianxiong Xiao.
\newblock Sun rgb-d: A rgb-d scene understanding benchmark suite.
\newblock In \emph{Proceedings of the IEEE conference on computer vision and pattern recognition}, pages 567--576, 2015.

\bibitem[Stojanov et~al.(2021)Stojanov, Thai, and Rehg]{stojanov2021using}
Stefan Stojanov, Anh Thai, and James~M Rehg.
\newblock Using shape to categorize: Low-shot learning with an explicit shape bias.
\newblock In \emph{Proceedings of the IEEE/CVF conference on computer vision and pattern recognition}, pages 1798--1808, 2021.

\bibitem[Sun et~al.(2024)Sun, Zhuang, Jiang, Liu, Xie, and Chandraker]{sun2024lidarf}
Shanlin Sun, Bingbing Zhuang, Ziyu Jiang, Buyu Liu, Xiaohui Xie, and Manmohan Chandraker.
\newblock Lidarf: Delving into lidar for neural radiance field on street scenes.
\newblock In \emph{Proceedings of the IEEE/CVF Conference on Computer Vision and Pattern Recognition}, pages 19563--19572, 2024.

\bibitem[Tang et~al.(2024)Tang, Han, Li, Yu, Hao, Hu, and Chen]{tang2024minigpt}
Yuan Tang, Xu Han, Xianzhi Li, Qiao Yu, Yixue Hao, Long Hu, and Min Chen.
\newblock Minigpt-3d: Efficiently aligning 3d point clouds with large language models using 2d priors.
\newblock \emph{arXiv preprint arXiv:2405.01413}, 2024.

\bibitem[Team et~al.(2024)Team, Ghosh, Walke, Pertsch, Black, Mees, Dasari, Hejna, Kreiman, Xu, et~al.]{team2024octo}
Octo~Model Team, Dibya Ghosh, Homer Walke, Karl Pertsch, Kevin Black, Oier Mees, Sudeep Dasari, Joey Hejna, Tobias Kreiman, Charles Xu, et~al.
\newblock Octo: An open-source generalist robot policy.
\newblock \emph{arXiv preprint arXiv:2405.12213}, 2024.

\bibitem[Tobin et~al.(2017)Tobin, Fong, Ray, Schneider, Zaremba, and Abbeel]{tobin2017domain}
Josh Tobin, Rachel Fong, Alex Ray, Jonas Schneider, Wojciech Zaremba, and Pieter Abbeel.
\newblock Domain randomization for transferring deep neural networks from simulation to the real world.
\newblock In \emph{2017 IEEE/RSJ international conference on intelligent robots and systems (IROS)}, pages 23--30. IEEE, 2017.

\bibitem[Uy et~al.(2019)Uy, Pham, Hua, Nguyen, and Yeung]{uy2019revisiting}
Mikaela~Angelina Uy, Quang-Hieu Pham, Binh-Son Hua, Thanh Nguyen, and Sai-Kit Yeung.
\newblock Revisiting point cloud classification: A new benchmark dataset and classification model on real-world data.
\newblock In \emph{Proceedings of the IEEE/CVF international conference on computer vision}, pages 1588--1597, 2019.

\bibitem[Vishwanath et~al.(2009)Vishwanath, Gupta, Vahdat, and Yocum]{vishwanath2009modelnet}
Kashi~Venkatesh Vishwanath, Diwaker Gupta, Amin Vahdat, and Ken Yocum.
\newblock Modelnet: Towards a datacenter emulation environment.
\newblock In \emph{2009 IEEE Ninth International Conference on Peer-to-Peer Computing}, pages 81--82. IEEE, 2009.

\bibitem[Wang et~al.(2023)Wang, Tang, Ji, Sun, Zhang, Ma, Zhao, Li, Zhao, Lv, et~al.]{wang2023beyond}
Haowei Wang, Jiji Tang, Jiayi Ji, Xiaoshuai Sun, Rongsheng Zhang, Yiwei Ma, Minda Zhao, Lincheng Li, Zeng Zhao, Tangjie Lv, et~al.
\newblock Beyond first impressions: Integrating joint multi-modal cues for comprehensive 3d representation.
\newblock In \emph{Proceedings of the 31st ACM International Conference on Multimedia}, pages 3403--3414, 2023.

\bibitem[Wang et~al.(2024)Wang, Xu, Dai, Xiang, Deng, Tong, and Yang]{wang2024moge}
Ruicheng Wang, Sicheng Xu, Cassie Dai, Jianfeng Xiang, Yu Deng, Xin Tong, and Jiaolong Yang.
\newblock Moge: Unlocking accurate monocular geometry estimation for open-domain images with optimal training supervision.
\newblock \emph{arXiv preprint arXiv:2410.19115}, 2024.

\bibitem[Wu et~al.(2023)Wu, Zhang, Fu, Wang, Ren, Pan, Wu, Yang, Wang, Qian, et~al.]{wu2023omniobject3d}
Tong Wu, Jiarui Zhang, Xiao Fu, Yuxin Wang, Jiawei Ren, Liang Pan, Wayne Wu, Lei Yang, Jiaqi Wang, Chen Qian, et~al.
\newblock Omniobject3d: Large-vocabulary 3d object dataset for realistic perception, reconstruction and generation.
\newblock In \emph{Proceedings of the IEEE/CVF Conference on Computer Vision and Pattern Recognition}, pages 803--814, 2023.

\bibitem[Wu et~al.(2022)Wu, Lao, Jiang, Liu, and Zhao]{wu2022point}
Xiaoyang Wu, Yixing Lao, Li Jiang, Xihui Liu, and Hengshuang Zhao.
\newblock Point transformer v2: Grouped vector attention and partition-based pooling.
\newblock \emph{Advances in Neural Information Processing Systems}, 35:\penalty0 33330--33342, 2022.

\bibitem[Xu et~al.(2024{\natexlab{a}})Xu, Cheng, Gao, Wang, Gao, and Shan]{xu2024instantmesh}
Jiale Xu, Weihao Cheng, Yiming Gao, Xintao Wang, Shenghua Gao, and Ying Shan.
\newblock Instantmesh: Efficient 3d mesh generation from a single image with sparse-view large reconstruction models.
\newblock \emph{arXiv preprint arXiv:2404.07191}, 2024{\natexlab{a}}.

\bibitem[Xu et~al.(2024{\natexlab{b}})Xu, Wang, Wang, Chen, Pang, and Lin]{xu2024pointllm}
Runsen Xu, Xiaolong Wang, Tai Wang, Yilun Chen, Jiangmiao Pang, and Dahua Lin.
\newblock Pointllm: Empowering large language models to understand point clouds.
\newblock In \emph{ECCV}, 2024{\natexlab{b}}.

\bibitem[Xu et~al.(2024{\natexlab{c}})Xu, Shi, Yifan, Chen, Yang, Peng, Shen, and Wetzstein]{xu2024grm}
Yinghao Xu, Zifan Shi, Wang Yifan, Hansheng Chen, Ceyuan Yang, Sida Peng, Yujun Shen, and Gordon Wetzstein.
\newblock Grm: Large gaussian reconstruction model for efficient 3d reconstruction and generation.
\newblock In \emph{European Conference on Computer Vision}, pages 1--20. Springer, 2024{\natexlab{c}}.

\bibitem[Yang et~al.(2024{\natexlab{a}})Yang, Yang, Hui, Zheng, Yu, Zhou, Li, Li, Liu, Huang, Dong, Wei, Lin, Tang, Wang, Yang, Tu, Zhang, Ma, Xu, Zhou, Bai, He, Lin, Dang, Lu, Chen, Yang, Li, Xue, Ni, Zhang, Wang, Peng, Men, Gao, Lin, Wang, Bai, Tan, Zhu, Li, Liu, Ge, Deng, Zhou, Ren, Zhang, Wei, Ren, Fan, Yao, Zhang, Wan, Chu, Liu, Cui, Zhang, and Fan]{qwen2}
An Yang, Baosong Yang, Binyuan Hui, Bo Zheng, Bowen Yu, Chang Zhou, Chengpeng Li, Chengyuan Li, Dayiheng Liu, Fei Huang, Guanting Dong, Haoran Wei, Huan Lin, Jialong Tang, Jialin Wang, Jian Yang, Jianhong Tu, Jianwei Zhang, Jianxin Ma, Jin Xu, Jingren Zhou, Jinze Bai, Jinzheng He, Junyang Lin, Kai Dang, Keming Lu, Keqin Chen, Kexin Yang, Mei Li, Mingfeng Xue, Na Ni, Pei Zhang, Peng Wang, Ru Peng, Rui Men, Ruize Gao, Runji Lin, Shijie Wang, Shuai Bai, Sinan Tan, Tianhang Zhu, Tianhao Li, Tianyu Liu, Wenbin Ge, Xiaodong Deng, Xiaohuan Zhou, Xingzhang Ren, Xinyu Zhang, Xipin Wei, Xuancheng Ren, Yang Fan, Yang Yao, Yichang Zhang, Yu Wan, Yunfei Chu, Yuqiong Liu, Zeyu Cui, Zhenru Zhang, and Zhihao Fan.
\newblock Qwen2 technical report.
\newblock \emph{arXiv preprint arXiv:2407.10671}, 2024{\natexlab{a}}.

\bibitem[Yang et~al.(2023{\natexlab{a}})Yang, Chen, Qian, Madaan, Iyengar, Fouhey, and Chai]{yang2023llmgrounder}
Jianing Yang, Xuweiyi Chen, Shengyi Qian, Nikhil Madaan, Madhavan Iyengar, David~F. Fouhey, and Joyce Chai.
\newblock Llm-grounder: Open-vocabulary 3d visual grounding with large language model as an agent, 2023{\natexlab{a}}.

\bibitem[Yang et~al.(2024{\natexlab{b}})Yang, Yang, Gupta, Han, Fei-Fei, and Xie]{yang2024thinking}
Jihan Yang, Shusheng Yang, Anjali~W Gupta, Rilyn Han, Li Fei-Fei, and Saining Xie.
\newblock Thinking in space: How multimodal large language models see, remember, and recall spaces.
\newblock \emph{arXiv preprint arXiv:2412.14171}, 2024{\natexlab{b}}.

\bibitem[Yang et~al.(2024{\natexlab{c}})Yang, Kang, Huang, Xu, Feng, and Zhao]{yang2024depth}
Lihe Yang, Bingyi Kang, Zilong Huang, Xiaogang Xu, Jiashi Feng, and Hengshuang Zhao.
\newblock Depth anything: Unleashing the power of large-scale unlabeled data.
\newblock In \emph{Proceedings of the IEEE/CVF Conference on Computer Vision and Pattern Recognition}, pages 10371--10381, 2024{\natexlab{c}}.

\bibitem[Yang et~al.(2024{\natexlab{d}})Yang, Kang, Huang, Zhao, Xu, Feng, and Zhao]{yang2024depthv2}
Lihe Yang, Bingyi Kang, Zilong Huang, Zhen Zhao, Xiaogang Xu, Jiashi Feng, and Hengshuang Zhao.
\newblock Depth anything v2.
\newblock \emph{arXiv preprint arXiv:2406.09414}, 2024{\natexlab{d}}.

\bibitem[Yang et~al.(2023{\natexlab{b}})Yang, Liu, Zhang, Pan, Guo, Li, Chen, Gao, Guo, and Zhang]{yang2023lidar}
Senqiao Yang, Jiaming Liu, Ray Zhang, Mingjie Pan, Zoey Guo, Xiaoqi Li, Zehui Chen, Peng Gao, Yandong Guo, and Shanghang Zhang.
\newblock Lidar-llm: Exploring the potential of large language models for 3d lidar understanding.
\newblock \emph{arXiv preprint arXiv:2312.14074}, 2023{\natexlab{b}}.

\bibitem[Yang et~al.(2023{\natexlab{c}})Yang, Guo, Xiong, Liu, Pan, Wang, Tong, and Guo]{yang2023swin3dpretrainedtransformerbackbone}
Yu-Qi Yang, Yu-Xiao Guo, Jian-Yu Xiong, Yang Liu, Hao Pan, Peng-Shuai Wang, Xin Tong, and Baining Guo.
\newblock Swin3d: A pretrained transformer backbone for 3d indoor scene understanding, 2023{\natexlab{c}}.

\bibitem[Yao et~al.(2020)Yao, Luo, Li, Zhang, Ren, Zhou, Fang, and Quan]{yao2020blendedmvs}
Yao Yao, Zixin Luo, Shiwei Li, Jingyang Zhang, Yufan Ren, Lei Zhou, Tian Fang, and Long Quan.
\newblock Blendedmvs: A large-scale dataset for generalized multi-view stereo networks.
\newblock In \emph{Proceedings of the IEEE/CVF conference on computer vision and pattern recognition}, pages 1790--1799, 2020.

\bibitem[Yeshwanth et~al.(2023)Yeshwanth, Liu, Nie{\ss}ner, and Dai]{yeshwanth2023scannet++}
Chandan Yeshwanth, Yueh-Cheng Liu, Matthias Nie{\ss}ner, and Angela Dai.
\newblock Scannet++: A high-fidelity dataset of 3d indoor scenes.
\newblock In \emph{Proceedings of the IEEE/CVF International Conference on Computer Vision}, pages 12--22, 2023.

\bibitem[Yin et~al.(2023)Yin, Zhang, Chen, Cai, Yu, Wang, Chen, and Shen]{yin2023metric3d}
Wei Yin, Chi Zhang, Hao Chen, Zhipeng Cai, Gang Yu, Kaixuan Wang, Xiaozhi Chen, and Chunhua Shen.
\newblock Metric3d: Towards zero-shot metric 3d prediction from a single image.
\newblock In \emph{Proceedings of the IEEE/CVF International Conference on Computer Vision}, pages 9043--9053, 2023.

\bibitem[Yin et~al.(2024)Yin, Liu, Xiao, Cohen-Or, Huang, and Chen]{yin2024sai3d}
Yingda Yin, Yuzheng Liu, Yang Xiao, Daniel Cohen-Or, Jingwei Huang, and Baoquan Chen.
\newblock Sai3d: Segment any instance in 3d scenes.
\newblock In \emph{Proceedings of the IEEE/CVF Conference on Computer Vision and Pattern Recognition}, pages 3292--3302, 2024.

\bibitem[Yu et~al.(2024)Yu, Duan, Herrmann, Freeman, and Wu]{yu2024wonderworld}
Hong-Xing Yu, Haoyi Duan, Charles Herrmann, William~T Freeman, and Jiajun Wu.
\newblock Wonderworld: Interactive 3d scene generation from a single image.
\newblock \emph{arXiv preprint arXiv:2406.09394}, 2024.

\bibitem[Yu et~al.(2016)Yu, Poirson, Yang, Berg, and Berg]{yu2016modeling}
Licheng Yu, Patrick Poirson, Shan Yang, Alexander~C Berg, and Tamara~L Berg.
\newblock Modeling context in referring expressions.
\newblock In \emph{Computer Vision--ECCV 2016: 14th European Conference, Amsterdam, The Netherlands, October 11-14, 2016, Proceedings, Part II 14}, pages 69--85. Springer, 2016.

\bibitem[Yu et~al.(2019)Yu, Yu, Cui, Tao, and Tian]{yu2019deep}
Zhou Yu, Jun Yu, Yuhao Cui, Dacheng Tao, and Qi Tian.
\newblock Deep modular co-attention networks for visual question answering.
\newblock In \emph{Proceedings of the IEEE/CVF conference on computer vision and pattern recognition}, pages 6281--6290, 2019.

\bibitem[Yuan et~al.(2024)Yuan, Ren, Feng, Zhao, Cui, and Li]{yuan2024visual}
Zhihao Yuan, Jinke Ren, Chun-Mei Feng, Hengshuang Zhao, Shuguang Cui, and Zhen Li.
\newblock Visual programming for zero-shot open-vocabulary 3d visual grounding.
\newblock In \emph{Proceedings of the IEEE/CVF Conference on Computer Vision and Pattern Recognition}, pages 20623--20633, 2024.

\bibitem[Zhang et~al.(2022)Zhang, Luo, Yu, Cui, and Lu]{zhang2022accelerating}
Gongjie Zhang, Zhipeng Luo, Yingchen Yu, Kaiwen Cui, and Shijian Lu.
\newblock Accelerating {DETR} convergence via semantic-aligned matching.
\newblock In \emph{Proceedings of the IEEE/CVF conference on computer vision and pattern recognition}, pages 949--958, 2022.

\bibitem[Zhang et~al.(2023{\natexlab{a}})Zhang, Lin, Wu, Luo, Xue, Lu, Wang, et~al.]{zhang2023online}
Gongjie Zhang, Jiahao Lin, Shuang Wu, Zhipeng Luo, Yang Xue, Shijian Lu, Zuoguan Wang, et~al.
\newblock Online map vectorization for autonomous driving: A rasterization perspective.
\newblock \emph{Advances in Neural Information Processing Systems (NeurIPS)}, 36:\penalty0 31865--31877, 2023{\natexlab{a}}.

\bibitem[Zhang et~al.(2023{\natexlab{b}})Zhang, Luo, Cui, Lu, and Xing]{metadetr}
Gongjie Zhang, Zhipeng Luo, Kaiwen Cui, Shijian Lu, and Eric~P Xing.
\newblock Meta-detr: Image-level few-shot detection with inter-class correlation exploitation.
\newblock \emph{IEEE transactions on pattern analysis and machine intelligence}, 45\penalty0 (11):\penalty0 12832--12843, 2023{\natexlab{b}}.

\bibitem[Zhang et~al.(2024)Zhang, Luo, Huang, Lu, and Xing]{zhang2024semantic}
Gongjie Zhang, Zhipeng Luo, Jiaxing Huang, Shijian Lu, and Eric~P Xing.
\newblock Semantic-aligned matching for enhanced {DETR} convergence and multi-scale feature fusion.
\newblock \emph{International Journal of Computer Vision}, 132\penalty0 (8):\penalty0 2825--2844, 2024.

\bibitem[Zhang et~al.(2023{\natexlab{c}})Zhang, Roller, Goyal, Artetxe, Chen, Chen, Dewan, Diab, Li, Lin, et~al.]{zhang2023opt}
Susan Zhang, Stephen Roller, Naman Goyal, Mikel Artetxe, Moya Chen, Shuohui Chen, Christopher Dewan, Mona Diab, Xian Li, Xi~Victoria Lin, et~al.
\newblock Opt: Open pre-trained transformer language models, 2022.
\newblock \emph{URL https://arxiv. org/abs/2205.01068}, 3:\penalty0 19--0, 2023{\natexlab{c}}.

\bibitem[Zhang et~al.(2023{\natexlab{d}})Zhang, Gong, and Chang]{zhang2023multi3drefer}
Yiming Zhang, ZeMing Gong, and Angel~X Chang.
\newblock Multi3drefer: Grounding text description to multiple 3d objects.
\newblock In \emph{Proceedings of the IEEE/CVF International Conference on Computer Vision}, pages 15225--15236, 2023{\natexlab{d}}.

\bibitem[Zheng et~al.(2024{\natexlab{a}})Zheng, Huang, and Wang]{zheng2024video3dllmlearningpositionaware}
Duo Zheng, Shijia Huang, and Liwei Wang.
\newblock Video-3d llm: Learning position-aware video representation for 3d scene understanding, 2024{\natexlab{a}}.

\bibitem[Zheng et~al.(2020)Zheng, Zhang, Li, Tang, Gao, and Zhou]{Structured3D}
Jia Zheng, Junfei Zhang, Jing Li, Rui Tang, Shenghua Gao, and Zihan Zhou.
\newblock Structured3d: A large photo-realistic dataset for structured 3d modeling.
\newblock In \emph{Proceedings of The European Conference on Computer Vision (ECCV)}, 2020.

\bibitem[Zheng et~al.(2024{\natexlab{b}})Zheng, Chiang, Sheng, Zhuang, Wu, Zhuang, Lin, Li, Li, Xing, et~al.]{zheng2024judging}
Lianmin Zheng, Wei-Lin Chiang, Ying Sheng, Siyuan Zhuang, Zhanghao Wu, Yonghao Zhuang, Zi Lin, Zhuohan Li, Dacheng Li, Eric Xing, et~al.
\newblock Judging llm-as-a-judge with mt-bench and chatbot arena.
\newblock \emph{Advances in Neural Information Processing Systems}, 36, 2024{\natexlab{b}}.

\bibitem[Zheng et~al.(2024{\natexlab{c}})Zheng, Li, Nagano, Liu, Hilliges, and De~Mello]{zheng2024unified}
Yufeng Zheng, Xueting Li, Koki Nagano, Sifei Liu, Otmar Hilliges, and Shalini De~Mello.
\newblock A unified approach for text-and image-guided 4d scene generation.
\newblock In \emph{Proceedings of the IEEE/CVF Conference on Computer Vision and Pattern Recognition}, pages 7300--7309, 2024{\natexlab{c}}.

\bibitem[Zhou et~al.(2024{\natexlab{a}})Zhou, Cheng, Yu, Tian, and Yuan]{zhou2024holodreamer}
Haiyang Zhou, Xinhua Cheng, Wangbo Yu, Yonghong Tian, and Li Yuan.
\newblock Holodreamer: Holistic 3d panoramic world generation from text descriptions.
\newblock \emph{arXiv preprint arXiv:2407.15187}, 2024{\natexlab{a}}.

\bibitem[Zhou et~al.(2023)Zhou, Wang, Ma, Liu, Huang, and Wang]{zhou2023uni3d}
Junsheng Zhou, Jinsheng Wang, Baorui Ma, Yu-Shen Liu, Tiejun Huang, and Xinlong Wang.
\newblock Uni3d: Exploring unified 3d representation at scale.
\newblock \emph{arXiv preprint arXiv:2310.06773}, 2023.

\bibitem[Zhou et~al.(2025)Zhou, Zhang, and Liu]{zhou2025diffgs}
Junsheng Zhou, Weiqi Zhang, and Yu-Shen Liu.
\newblock Diffgs: Functional gaussian splatting diffusion.
\newblock \emph{Advances in Neural Information Processing Systems}, 37:\penalty0 37535--37560, 2025.

\bibitem[Zhou et~al.(2024{\natexlab{b}})Zhou, Fan, Xu, Chang, Chari, Bharadwaj, You, Wang, and Kadambi]{zhou2024dreamscene360}
Shijie Zhou, Zhiwen Fan, Dejia Xu, Haoran Chang, Pradyumna Chari, Tejas Bharadwaj, Suya You, Zhangyang Wang, and Achuta Kadambi.
\newblock Dreamscene360: Unconstrained text-to-3d scene generation with panoramic gaussian splatting.
\newblock In \emph{European Conference on Computer Vision}, pages 324--342. Springer, 2024{\natexlab{b}}.

\bibitem[Zhu et~al.(2024)Zhu, Wang, Zhang, Pang, and Liu]{zhu2024llava}
Chenming Zhu, Tai Wang, Wenwei Zhang, Jiangmiao Pang, and Xihui Liu.
\newblock Llava-3d: A simple yet effective pathway to empowering lmms with 3d-awareness.
\newblock \emph{arXiv preprint arXiv:2409.18125}, 2024.

\bibitem[Zhu et~al.(2023{\natexlab{a}})Zhu, Kumar, Hu, and Liu]{zhu2023tame}
Shengjie Zhu, Abhinav Kumar, Masa Hu, and Xiaoming Liu.
\newblock Tame a wild camera: In-the-wild monocular camera calibration.
\newblock In \emph{NeurIPS}, 2023{\natexlab{a}}.

\bibitem[Zhu et~al.(2023{\natexlab{b}})Zhu, Ma, Chen, Deng, Huang, and Li]{zhu20233d}
Ziyu Zhu, Xiaojian Ma, Yixin Chen, Zhidong Deng, Siyuan Huang, and Qing Li.
\newblock 3d-vista: Pre-trained transformer for 3d vision and text alignment.
\newblock In \emph{Proceedings of the IEEE/CVF International Conference on Computer Vision}, pages 2911--2921, 2023{\natexlab{b}}.

\end{thebibliography}
}

\clearpage
\setcounter{page}{1}
\maketitlesupplementary
\section{Appendix}
\begin{figure*}
    \centering
    \includegraphics[width=\linewidth]{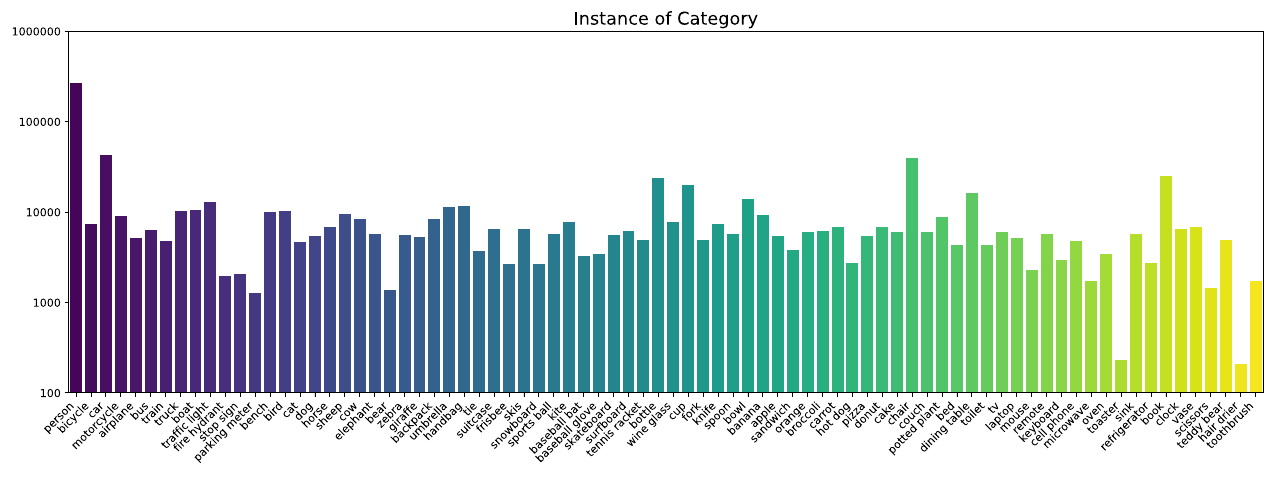}
    \caption{\textbf{Statistic of COCO-3D.} The number of instances for each category.}
    \label{fig:stat_category_instance}
\end{figure*}

\begin{figure*}
    \centering
    \includegraphics[width=\linewidth]{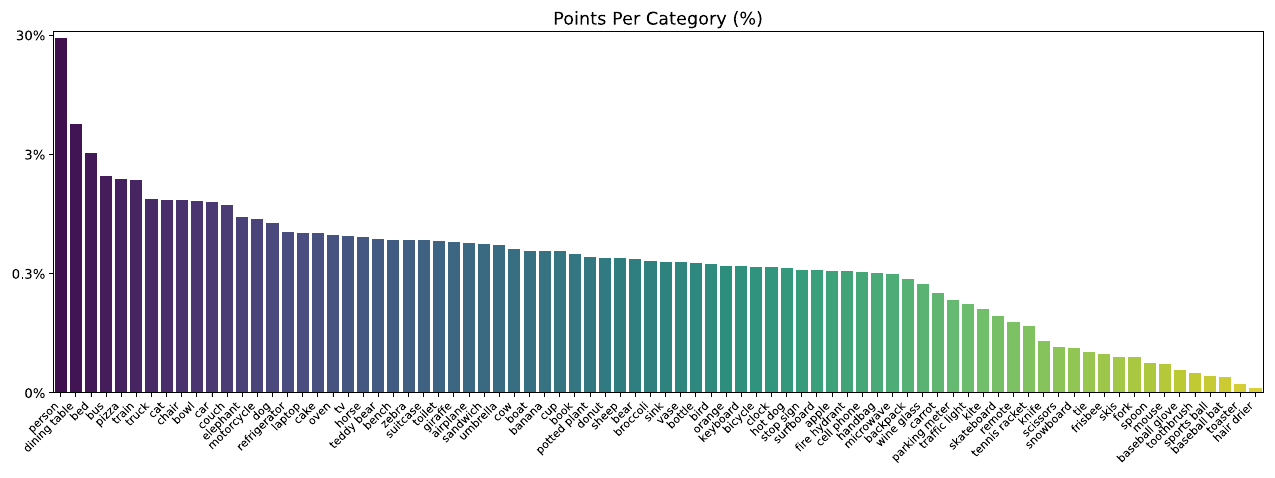}
    \caption{\textbf{Statistic of COCO-3D.} The percentage distribution of number points across various categories.}
    \label{fig:stat_category_point}
\end{figure*}

\paragraph{Statistics of COCO-3D}
\cref{fig:stat_category_instance} shows the number of instances for each category. The x-axis lists the categories, while the y-axis represents the instance count. \cref{fig:stat_category_point} illustrates the percentage distribution of points across different categories. The x-axis represents the various categories, and the y-axis indicates the percentage of points assigned to each category. From the figures, it is evident that most points are concentrated in the ``person" category, which accounts for 30\% of the total points—far exceeding the other categories. Compared to other domain-specific 3D datasets, our dataset exhibits notable differences. COCO-3D is derived from the transformation of COCO data, which enables us to retain the rich semantic information and diverse annotations found in COCO. Our experiments have demonstrated that our synthetic data performs well in zero-shot transfer, giving us confidence in leveraging this dataset to enhance 3D object detection and recognition. It is particularly worth mentioning that our dataset includes a large number of scenes involving people, with especially abundant data in the ``person" category. This makes our dataset more realistic when addressing human-related tasks. Pre-training on synthetic data followed by fine-tuning on real data can, to some extent, alleviate the challenges posed by the scarcity of real data.

\paragraph{Compare with Other 3D Datasets}
Compared to traditional databases \cref{tab:datasets} (such as ShapeNet \cite{chang2015shapenet}, ModelNet \cite{vishwanath2009modelnet}, 3D-Future \cite{fu20213d} that mainly focus on single objects, ScanNet \cite{dai2017scannet}, Matterport3D \cite{Matterport3D} that are limited to small-scale scenes), or SUN-RGBD \cite{song2015sun} and Omni3D \cite{brazil2023omni3d} only include monocular 3D representation datasets of indoor scenes, our COCO-3D and object365-v2-3D datasets are significantly ahead in terms of the number of scenes and categories. Specifically, COCO-3D contains 122K scene instances and 81 categories, while object365-v2-3D has 2M scene instances and 365 categories. Our dataset includes indoor and outdoor scenes. Although the data is synthetic, rich experimental results prove that it has zero shot capabilities and can be generalized to other datasets, providing sufficient data support for tasks such as 3D perception.

\paragraph{Discussion with SpatialVLM}
SpatialVLM \cite{chen2024spatialvlm} improves the spatial QA performance of VLM by converting 2D images into 3D point clouds and generating many spatial QA pairs. However, it does not calibrate the point cloud’s geometric accuracy or camera parameters, nor does it carry out systematic validation on low-level 3D vision tasks such as segmentation, etc. It only addresses QA tasks about relative positions and sizes of objects. In contrast, our work builds a full 3D representation, of which the point cloud is only one part. For each scene, we calibrate gravity direction, camera parameters, and metric scale. Moreover, our experiments cover a range of spatial reasoning tasks, from low-level (semantic segmentation, instance segmentation, few-shot learning, zero-shot learning) to high-level (QA, captioning, and referring segmentation). 

\begin{table}
    \renewcommand{\arraystretch}{1.3}
    \setlength{\tabcolsep}{5pt}
    \centering
    \resizebox{\linewidth}{!}{%
        \begin{tabular}{lcccc}
        \toprule
        Dataset            & Number      & Categories & Class & Scenes/Objects \\ \midrule
        ShapeNet \cite{chang2015shapenet}           & 51k         & 55         &  -    & Objects        \\
        ModelNet  \cite{vishwanath2009modelnet}         & 12k         & 40         &  -    & Objects        \\
        3D-Future  \cite{fu20213d}        & 16k         & 34         &  -    & Objects        \\
        ABO \cite{collins2022abo}                & 8k          & 63         &  -    & Objects        \\
        Toys4K  \cite{stojanov2021using}           & 4k          & 105        &  -    & Objects        \\
        CO3D V1 / V2 \cite{reizenstein2021commonobjects3dlargescale}      & 19 / 40k    & 50         &  -    & Objects        \\
        ScanObjectNN \cite{uy2019revisiting}      & 15k         & 15         &  -    & Objects        \\
        GSO \cite{downs2022google}               & 1k          & 17         &  -    & Objects        \\
        AKB-48 \cite{liu2022akb}            & 2k          & 48         &  -    & Objects        \\
        OmniObject3D \cite{wu2023omniobject3d}      & 6k          & 190        &  -    & Objects        \\ \midrule
        LLFF \cite{mildenhall2019llff}              & 35          &  -         &  -    & Scenes         \\
        DTU \cite{aanaes2016large}               & 124         &  -         &  -    & Scenes         \\
        BlendedMVS \cite{yao2020blendedmvs}        & 133         &  -         &  -    & Scenes         \\
        ScanNet \cite{dai2017scannet}          & 1509        &  -         & 20    & Scenes         \\
        Matterport3D \cite{Matterport3D}      & 90          &  -         & 21    & Scenes         \\
        Tanks and Temples \cite{Knapitsch2017}  & 21          &  -         &  -    & Scenes         \\
        ETH3D \cite{schoeps2017cvpr}             & 25          &  -         &  -    & Scenes         \\
        ARKitScenes \cite{baruch2021arkitscenes}       & 1004        &  -         &  -    & Scenes         \\
        ScanNet++ \cite{yeshwanth2023scannet++}         & 460         &  -         & 100   & Scenes         \\
        S3DIS \cite{armeni20163d}             & 271         &  -         & 13    & Scenes         \\
        Structured3D \cite{Structured3D}      & 3500        &  -         & 25    & Scenes         \\ 
        \midrule
        COCO-3D            & 122K     &  -         & 81    & Scenes         \\
        object365-v2-3D    & 2M        &  -         & 365    & Scenes         \\
        \bottomrule
        \end{tabular}
    }
\label{tab:datasets}
\caption{\textbf{A comparison between COCO-3D, Object365-v2-3D, and other commonly-used 3D scenes/object datasets.}}
\end{table}

\begin{figure*}
    \centering
    \includegraphics[width=\linewidth]{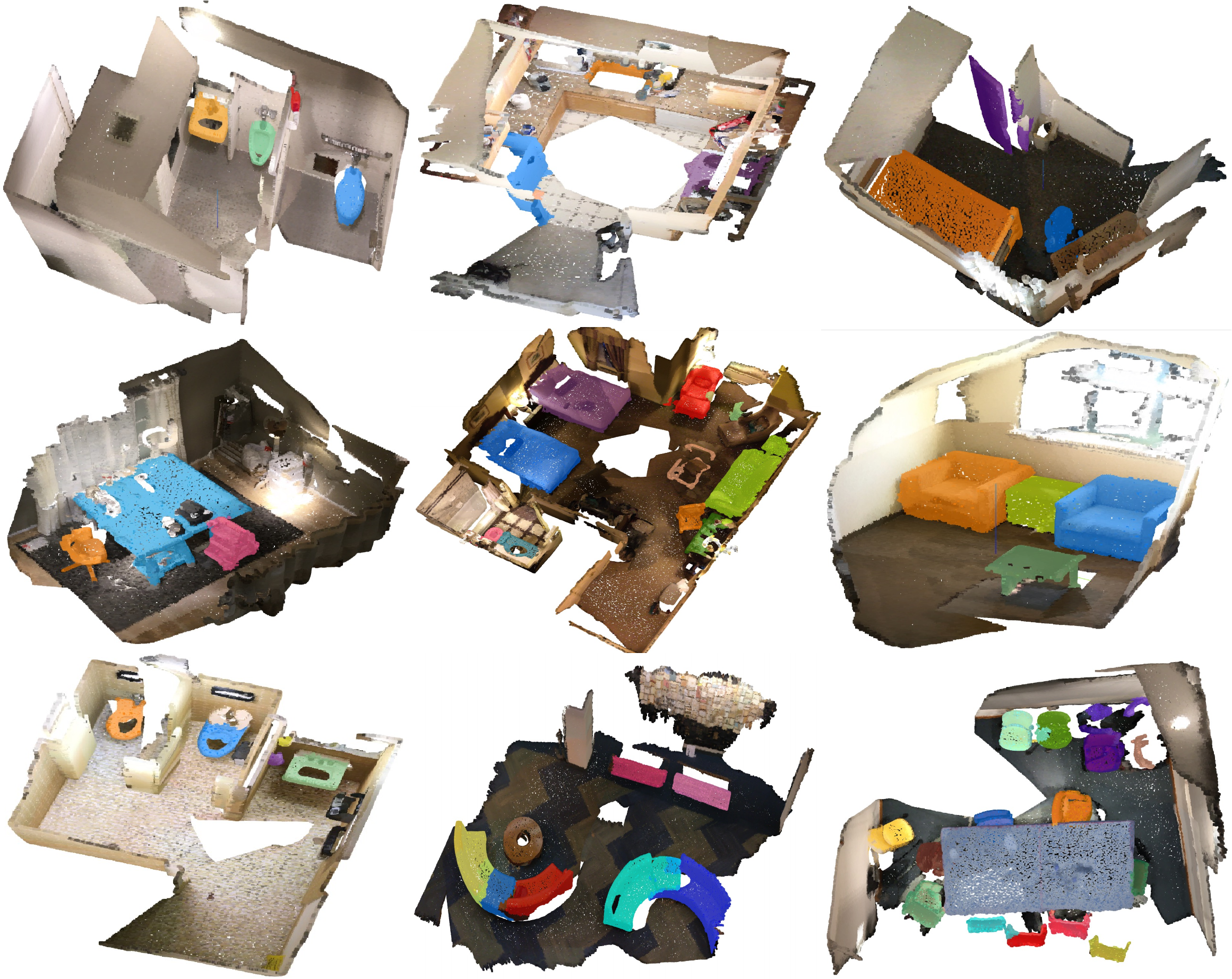}
    \caption{\textbf{Visualization of zero-shot point cloud instance segmentation results.} Despite significant differences between synthetic and real data, models trained on COCO-3D can directly generalize to ScanNet.}
    \label{fig:supp_ins_vis}
\end{figure*}

\begin{figure*}
    \centering
    \includegraphics[width=\linewidth]{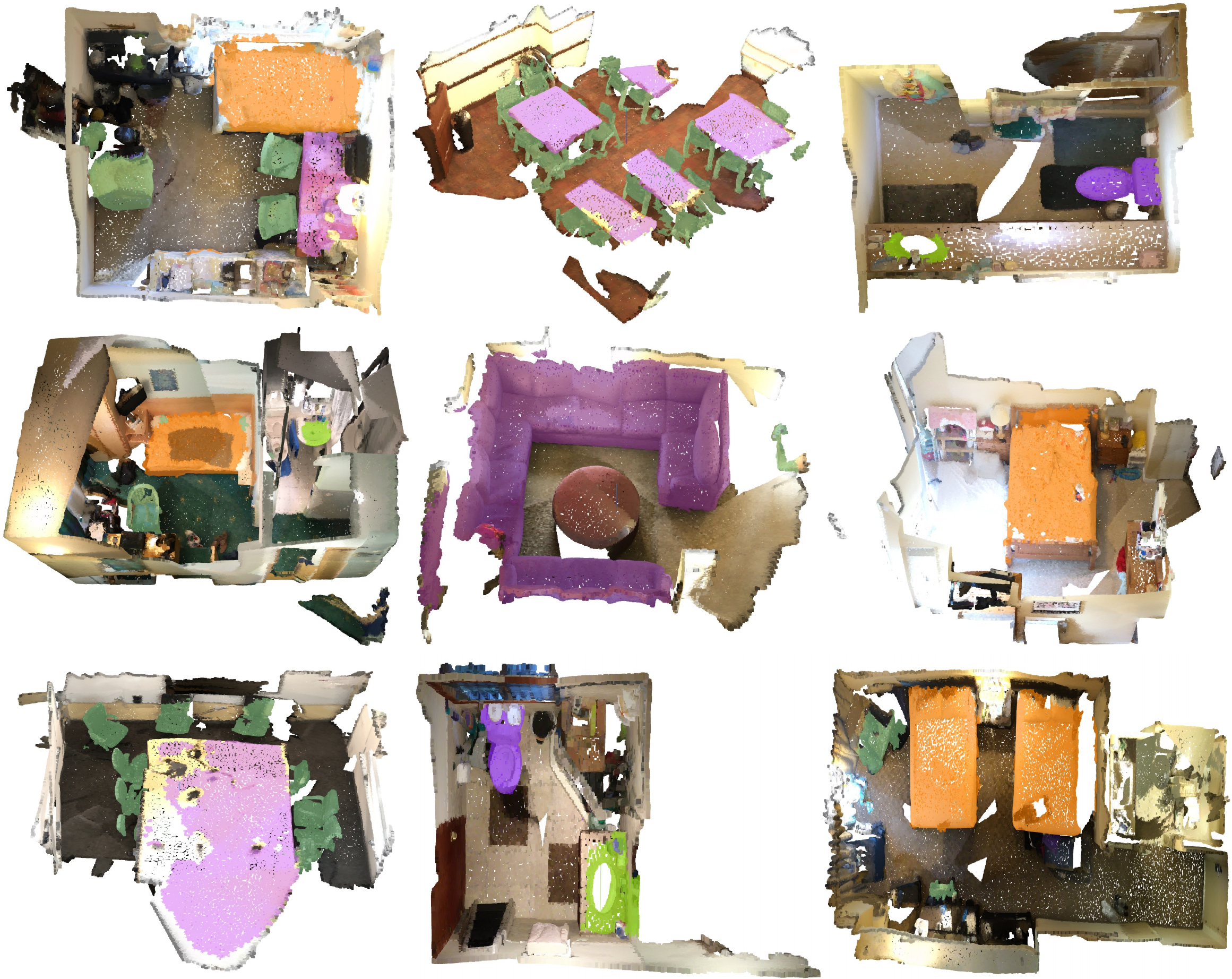}
    \caption{\textbf{Visualization of zero-shot point cloud semantic segmentation results.} Despite significant differences between synthetic and real data, models trained on COCO-3D can directly generalize to ScanNet.}
    \label{fig:supp_sem_vis}
\end{figure*}

\paragraph{More Visualization}
In \cref{fig:supp_ins_vis} and \cref{fig:supp_sem_vis}, we provide more visualization results of the zero-shot experiments on ScanNet for Uni3D.

\paragraph{Data Quality Assurance}
In the process of constructing the dataset from 2D images to 3D representations, we implemented a series of data quality assurance mechanisms to ensure that the generated data meets high standards in terms of authenticity, accuracy, and consistency. First, through depth estimation and camera parameter prediction, we use an automatic filtering algorithm after generating a preliminary 3D representation to remove edge areas, undefined areas, and predicted abnormal points, and calculate the scale factor based on the relative depth and quantized depth distribution in the valid point set to achieve an effective fusion of depth information and absolute scale. Next, we select some samples and use Open3D visualization for manual verification to verify the consistency between the original 2D annotations and the generated 3D annotations, and check the correspondence between the 3D representation and the original 2D image, so as to promptly discover and correct possible errors in the automatic process. Finally, we further ensure the rationality of the data in scale and structure by statistically analyzing the size distribution of each category and comparing it with the actual physical size.

\end{document}